\documentclass[3p,times]{elsarticle}
\usepackage{amssymb}

\usepackage[figuresright]{rotating}

\usepackage{times}
\usepackage{epsfig}
\usepackage{graphicx}
\usepackage{amsmath}
\usepackage{amssymb}
\usepackage{xcolor}

\usepackage{booktabs}

\usepackage{bm}

\usepackage{mathrsfs}
\usepackage[ruled,vlined,lines numbered]{algorithm2e}
\usepackage{subfigure}
\usepackage{enumerate}
\usepackage{caption}
\usepackage{multirow} 
\usepackage{booktabs}
 
\newcommand{\tabincell}[2]{\begin{tabular}{@{}#1@{}}#2\end{tabular}}  
\def\red{\color{black}}

\begin{document}

\begin{frontmatter}




\title{Embedding Deep Networks into Visual Explanations}


\author{Zhongang Qi\textsuperscript{\rm 1,2,\ddag}, Saeed Khorram\textsuperscript{\rm 1}, Li Fuxin\textsuperscript{\rm 1}}

\address{\textsuperscript{\rm 1}{\em School of Electrical Engineering and Computer Science, Oregon State University}\\
\textsuperscript{\rm 2}{\em Applied Research Center (ARC), PCG, Tencent}\\
zhongangqi@tencent.com, \{khorrams,lif\}@oregonstate.edu}

\begin{abstract}
In this paper, we propose a novel Explanation Neural Network (XNN) to explain the predictions made by a deep network. The XNN works by learning a nonlinear embedding of a high-dimensional activation vector of a deep network layer into a low-dimensional explanation space while retaining faithfulness i.e., the original deep learning predictions can be constructed from the few concepts extracted by our explanation network. We then visualize such concepts for human to learn about the high-level concepts that the deep network is using to make decisions. We propose an algorithm called Sparse Reconstruction Autoencoder (SRAE) for learning the embedding to the explanation space. SRAE aims to reconstruct part of the original feature space while retaining faithfulness. A pull-away term is applied to SRAE to make the bases of the explanation space more orthogonal to each other.
A visualization system is then introduced for human understanding of the features in the explanation space. The proposed method is applied to explain CNN models in image classification tasks. We conducted a human study, which shows that the proposed approach outperforms single saliency map baselines, and improves human performance on a difficult classification tasks. Also, several novel metrics are introduced to evaluate the performance of explanations quantitatively without human involvement.
\end{abstract}

\begin{keyword}
Deep Neural Networks, Embedding, Visual Explanations.
{\let\thefootnote\relax\footnotetext{{\ddag The work is done when Zhongang Qi was working at Oregon State University}}}
\end{keyword}

\end{frontmatter}

\section{Introduction}
\label{sec:introduction}

Deep learning has made significant strides in recent years, surpassing human performance in many tasks, such as image classification~\cite{Krizhevsky2012,He2016residual}, go-playing~\cite{silver2016mastering}, and classification of medical images~\cite{esteva2017dermatologist}. 
However, the usage of deep learning in real applications still must overcome a trust barrier. Imagine scenarios with a doctor facing a deep learning prediction: this CT image indicates malignant cancer, or a pilot facing a prediction: make an emergency landing immediately. These predictions may be backed up with a claimed high accuracy on benchmarks, but it is human nature not to trust them unless we are \textit{convinced} that they are reasonable for each individual case. The lack of trust is worsened because of known cases where adversarial examples can fool deep learning to output wrong answers~\cite{Szegedy2013,Goodfellow2014}. In order to establish trust, humans need to understand how deep learning makes decisions. Such understanding could also help humans to gain additional insights into new problems, potentially improve deep learning algorithms, and improve human-machine collaboration.

Dictionaries often contain explanations of a concept in the form ``A is something because of B, C, and D", e.g. this is a bird because it has feathers, wings, and a beak. This type of explanation has two properties. Firstly, it is concise -- there is no need for hundred reasons adding up to define A. Secondly, it relies on B, C, and D, which are also high-level concepts. Both of these properties are often at odds with standard deep learning predictions -- which are combinations of outputs from thousands of neurons in dozens of layers. 

Approaches have been proposed to visualize each of the filters~\cite{zeiler2014v} and for humans to name them~\cite{bau2017network}, but it is difficult for these approaches to obtain a concise representation. 
On the other hand, many other approaches generate attention maps that backtrack a decision to important parts of the original image~\cite{simonyan2014deep,cao2015look,zhou2016learning,zhang16excitationBP,selvaraju2016grad}. These are often nice and quite informative, but they are limited to explaining individual images and do not provide high-level concepts that can be broadly applicable to \textit{all} images in a category simultaneously.

\begin{figure*}[t]
\begin{center}
\subfigure[] { \label{fig:IntrEx}
\includegraphics[width=0.54\linewidth]{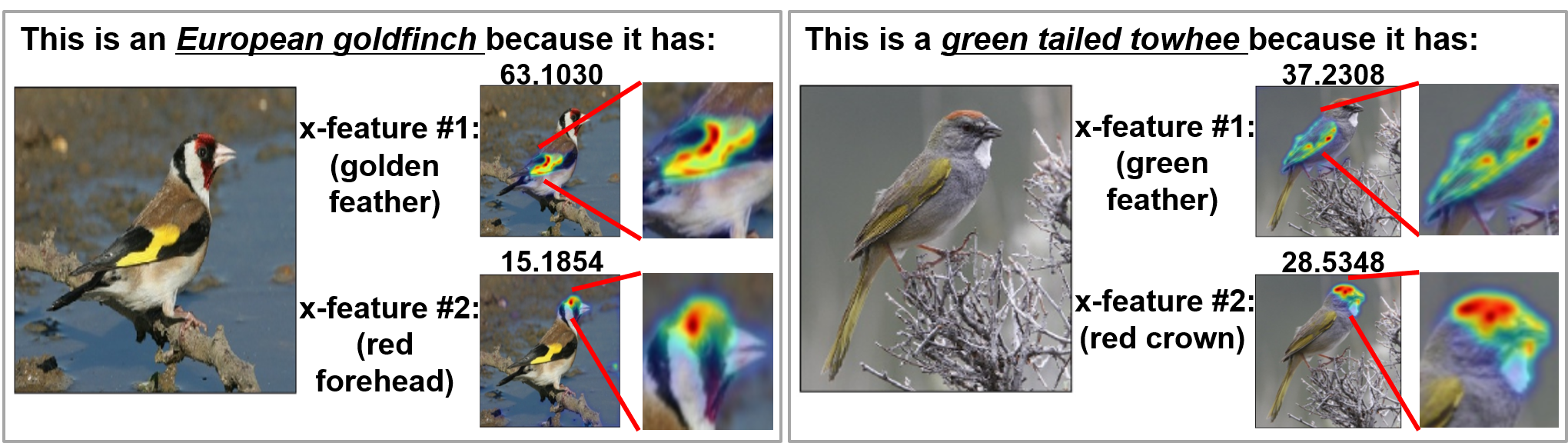}
}
\subfigure[] { \label{fig:XDL}
\includegraphics[width=0.41\linewidth]{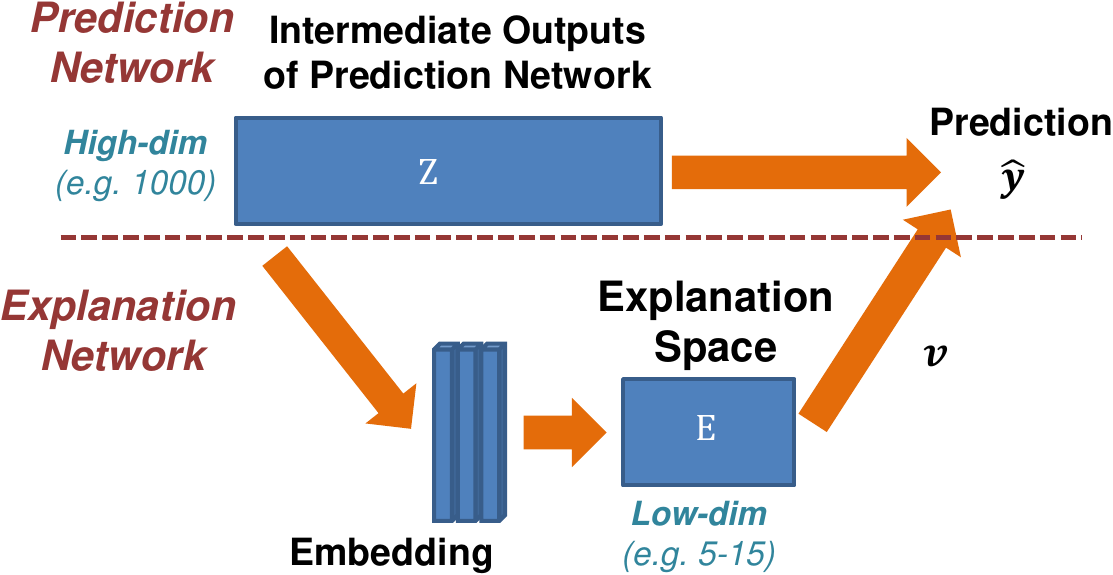}
}
\end{center}
\caption{\small (a) Example explanations that our approach can generate: the left figure shows that there are two key features (with their importance indicated on top of each heatmap) to predict {\em European goldfinch}: golden feather and red forehead. The right figure shows the two key features to predict {\em green tailed towhee}: green feather and red crown. Note that our approach generates the visualizations so that human can deduct those feature maps, it does not train/predict textual annotations; (b) Conceptually, the explanation network (XNN) mimics the original network with a nonlinear embedding layer so that the original deep learning prediction $\hat{\bf y}$ can be reproduced from a low-dimensional space. This \textit{Explanation Space} represents high-level features that are interpretable to humans. XNN can be attached to any layer of the network}
\label{fig:1}
\end{figure*}

In this paper, we attempt to make explanations similar to ``A is something because of B, C, D" by extracting several high-level concepts from deep networks to aid human understanding (Figure \ref{fig:IntrEx}). Our model 
attaches a separate explanation network (XNN) to a layer in the original deep network and reduces it to a few  concepts (named the \textit{Explanation Space}), from which one can generate predictions similar to the original deep network (Figure \ref{fig:XDL}). We also show that the visualizations of the concepts generated by our method are human understandable.


Our model directly infers concepts from the deep network. It does NOT train from ground truth concepts defined by human, either defined by labels, attributes, or text. The reason we deliberately choose not to use human concepts is to adapt to future situations where the deep network may perform a task in a domain in which human does not have expert knowledge. 
An example of that would be early-stage cancer prediction from medical imaging, where it may not be totally clear which patterns in the image may lead to cancer, but a deep network may be able to reliably find it out. XNN could then summarize these concepts and provide guidance to human experts and allow them to verify this tangible knowledge derived from the DNN and highlighted by XNN.

We evaluate our approach by 1) Human evaluation, where humans are presented different explanations to check which one improves their categorization capabilities. 2) Metric-based evaluation, where we define quantitative metrics for the aforementioned properties of an explanation network and evaluate them on two different datasets: a fine-grained bird classification and a scene recognition datasets which both have rich ground truth annotations allowing us to compute the introduced metrics. Although the experiments in the paper focus on convolutional neural networks (CNN) applied to images, the explanation framework we develop is general and applicable to other types of deep networks as well. 

This journal paper is an extension of our publication in a NeuRIPS 2017 workshop~\cite{QiXNN17}, which predates most other related approaches. We believe this is one of the first steps towards general explainable deep learning that can advance human knowledge and enhance future collaboration between humans and machines. In this journal paper, we improved the loss function and performed a comprehensive evaluation involving novel quantitative metrics as well as human studies.

Our contributions in this paper are as follows:
\begin{itemize}
\vspace{-0.07in}
\item We propose a novel explanation network to form a low-dimensional explainable concept space from deep networks. A sparse reconstruction auto-encoder with a pull-away term is proposed to make the explanation network faithful and orthogonal as defined previously.
\item We present a visualization paradigm for human understanding of the concept space.
\item We present a user study that shows our explanations can improve human performance on difficult tasks.
\item We propose automatic quantitative metrics to evaluate the performance of an explanation algorithm for faithfulness, locality and orthogonality.
Experimental results show that the proposed explanation methods provide insights to how the deep network models work.
\end{itemize}





\section{Model Formulation} \label{sec:mFormulation}
\subsection{The Explanation Network}
\label{sec:xdl}
Given a deep network (DNN) as a prediction model, we propose to learn an extra Explanation Neural Network (XNN) (Figure \ref{fig:XDL}), which can be attached to any intermediate layer of the DNN. 
The XNN attempts to learn an embedding to a low-dimensional explanation space, and then directly learns a mapping from the explanation space that mimics the output of the original DNN model. 
We denote the input feature space of the XNN as ${\bf Z}({\bf x};{\bf W})$, where ${\bf x}$ are the input features (in the case of CNNs, an image) and ${\bf W}$ are the parameters (from all layers) of the original DNN model, respectively. ${\bf Z}$ represents the output of an intermediate layer of the DNN.
The XNN is used to embed ${\bf Z}$ to an explanation space, denoted as ${\bf E}_{\bm{\theta}}({\bf Z})$, where $\bm{\theta}$ represents parameters of the embedding that need to be learned. As a shorthand, we will also refer to the explanation space as an \textit{x-layer}, and each dimension in the x-layer an \textit{x-feature}.
Note that during the explanation, we do not attempt to change the parameters ${\bf W}$ of the original DNN model. The explanation network can in principle be attached to any intermediate layer of the DNN, although the closer to the prediction, the higher level the concepts are and it becomes easier to mimic the prediction of the DNN with a low-dimensional embedding.

We believe that for the explanation network to be understandable, the number of x-features needs to be small since it is difficult for humans to parse numerous factors at the same time. 
In order to ensure this, we propose to explain binary classification or classification with a few outputs. An explanation to classification with hundreds of classes can be decomposed into explaining smaller tasks such as those mentioned.
For a prediction $\hat{\bf y}$ with lower dimensionality than the explanation space, we can safely assume that the explanation network could remain faithful to the prediction, since
a naive case would be to use $\hat{\bf y}$ as the explanation, which is perfectly faithful but not interpretable. Hence, the low-dimensional embedding ${\bf E}$ can also be thought of as expanding $\hat{\bf y}$ to a few more dimensions, therefore enriching the explanations. This is in contrast to some explanation methods that are built upon human concepts, which are more interpretable but lose a significant portion of the faithfulness to the original DNN (e.g.~\cite{zhou2018interpretable}). 

In this paper, we focus on attaching our proposed XNN to fully-connected layers, although some experiments attaching XNN to convolutional layers will also be shown. The concepts generated in these layers are high-level, and our conceptual goal is to visualize those concepts and to make humans learn them: human has an excellent deep neural network in the brain for learning and generalizing perceptual concepts. We would like to show humans examples from a small number of perceptual concepts from the explanation space so that people can utilize their own perceptual neural network for learning and naming those. Our primary tool for this display is heatmaps (e.g. Figure \ref{fig:IntrEx}) highlighting a specific region in the image, similar to those used in saliency models in prior work. Our work will provide several different and largely orthogonal concepts, visualized by heatmaps, for improving the understanding of the predictions from a DNN.
The two main topics in the explanation network are the embedding algorithm and the visualization of the explanations, which will be discussed in the next three subsections. 


\begin{figure*}[t]
\begin{center}
\subfigure[] { \label{fig:SAE}
\includegraphics[width=0.43\linewidth]{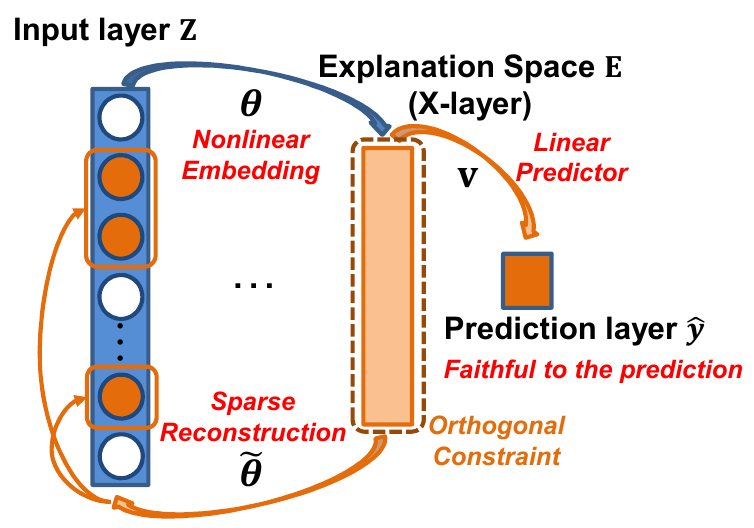}
}
\subfigure[] { \label{fig:Spa}
\includegraphics[width=0.25\linewidth]{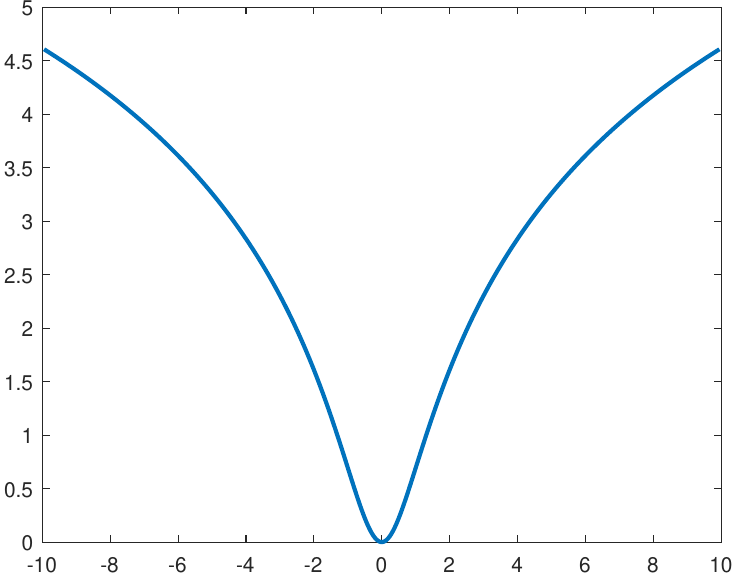}
}
\subfigure[] { \label{fig:Spb}
\includegraphics[width=0.24\linewidth]{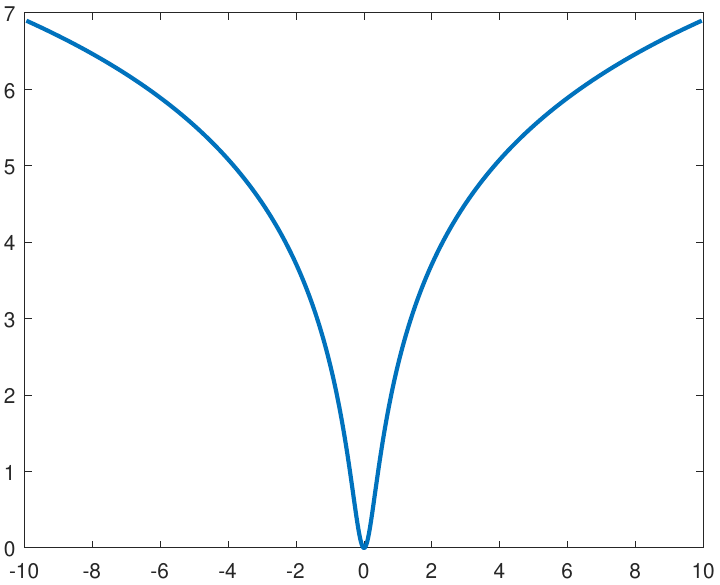}
}
\end{center}
\caption{(a) Illustration of the SRAE used for the explanation network. The
nonlinear embedding $\bm{\theta}$ embeds the input layer $\mathbf{Z}$ into the explanation
space $\mathbf{E}$, which is then used for predicting $\hat{\mathbf{y}}$; then from $\mathbf{E}$,
$\tilde{\bm{\theta}}$ performs sparse reconstruction on some dimensions of the
original input $\mathbf{Z}$. Also, the pull-away term enforces orthogonality constraints
on the network; (b) The log penalty function $\log (1+q\cdot r^{2})$ when $q=1$;
(c) The log penalty function $\log (1+q\cdot r^{2})$ when $q=10$.}
\label{fig:XDLOrth}
\end{figure*}


\subsection{Embedding to the explanation space}
\label{sec:FES}

We now explain the loss functions used in training the XNN piece-by-piece.
The first loss function in XNN attempts to be faithful to the prediction
of the original DNN:
%
\begin{align}
{L_{FS}} = \frac{1}{M}\sum _{i=1}^{M}L\Big {(}f_{\mathbf{v}}{\big (}{\mathbf{E}}_{
\bm{\theta }}(\mathbf{Z}^{(i)}){\big )}, \hat{\mathbf{y}}^{(i)}\Big {)}
\label{eq:basic}
\end{align}%
where the goal is to learn $({\mathbf v}, \bm{\theta})$, which maps DNN intermediate
output $\mathbf{Z}^{(i)}$ from training example $x^{(i)}$ first to the explanation
space $\left (\mathbf{E}_{\bm{\theta }}(\mathbf{Z}^{(i)})\right )$, and then to
the output space $\left (f_{\mathbf{v}}(\cdot )\right )$ to mimic the output
$\hat{\mathbf{y}}^{(i)}$ from the DNN on $x^{(i)}$. For
$\hat{\mathbf{y}}^{(i)}$, we usually use the DNN output in the layer before
the softmax layer (referred to in some papers as \textit{logits}) to prevent
interactions with predictions on other categories; $M$ is the number of
training examples; $L$ is a loss function, usually a regression loss such
as squared loss or log loss. However, as we argued in Sec.~\ref{sec:xdl},
if only this loss is minimized, we will obtain a degenerate solution by
using $\hat{\mathbf{y}}^{(i)}$ as the explanation, which will not bring any
insights. Hence, additional terms need to be added to prevent degeneracy
and improve interpretability.

We hypothesize that explanations may become more effective if they reconstruct
the original high-dimensional feature space better. Since by better reconstructing
$\textbf{Z}$ there need to be features that are richer than the simple output
$\hat{\mathbf{{y}}}$, and we would be able to locate several
\textit{diverse} $x$-features that simultaneously contribute to
$\hat{\mathbf{{y}}}$ and prevent degeneracy. However, the issue with a reconstruction
term is that features irrelevant to the prediction target may also be reconstructed,
since the original $\textbf{Z}$ is high-dimensional and a significant portion
of it may not be relevant to the prediction $\hat{\mathbf{{y}}}$ on one or a
few classes.

To avoid this, we propose to enhance the objective by adding a novel sparsity
term which reconstructs \textit{some dimensions} of the original features
$\mathbf{Z}$, but not all of them. By attempting to reconstruct some dimensions
of $\mathbf{Z}$ with only a few embeddings, and to mimic the original predictions
$\hat{\mathbf{y}}$ with the same embeddings, the maximal number of diverse information
that is relevant to $\hat{\mathbf{y}}$ in $\mathbf{Z}$ needs to be packed in the
low-dimensional space. Packing redundant information in correlated dimensions
would be harmful for reconstruction, and reconstructing irrelevant features
would harm the ability to recover $\hat{\mathbf{y}}$. By introducing a sparse
penalty, we define the sparse reconstruction loss as:
%
\begin{equation}
L_{SR} = \text{Sparsity}(\mathbf{Q});~~ Q_{k} = \frac{1}{M}\sum _{i=1}^{M}
L{\Big (}{\tilde{Z}_{k}^{(i)}, Z_{k}^{(i)}{\Big )}}
\label{eq:sparse}
\end{equation}
where $\tilde{Z}^{(i)}_k$ is the reconstruction of $Z_{k}^{(i)}$, and
$Q_{k}$, $Z_{k}^{(i)}$, and $\tilde{Z}^{(i)}_{k}$ are the $k$-th dimension
of $\mathbf{Q}$, $\mathbf{Z}^{(i)}$, and $\tilde{\mathbf{Z}}^{(i)}$, respectively.
In the optimization, $Q_{k}$ measures the capability of reconstructing
the $k$-th dimension of $\mathbf{Z}$.

The Sparsity$(\cdot )$ term could be any term inducing sparsity.
In our proposed model we choose the log penalty $\log(1 + q\cdot r^2)$ \cite{lee2007efficient} to achieve the sparsity of the reconstruction errors for different features, where $q>0$ is a sparsity parameter (Fig.~\ref{fig:XDLOrth}(b-c)). Here $r^2$
is the average squared reconstruction loss on each dimension over the whole training set, which equals to $Q_k$ using a square loss: $\text{Sparsity}(\mathbf{Q})=\frac{1}{S_z}\sum_{k=1}^{S_z}\log(1+q\cdot Q_k)$, $Q_k=\frac{1}{M}\sum_{i=1}^{M}||\tilde{Z}^{(i)}_k - Z^{(i)}_k||^2$, where $S_z$ is the dimensionality of $\mathbf{Z}$.
{The log penalty
(Fig.~\ref{fig:XDLOrth}(b-c)) is a robust loss function, in the sense
that large $r$ increases the loss function sublinearly (less than an
$L_{1}$ penalty $|r|$ where the increase is linear). Some dimensions of
$\textbf{Z}$ can afford to have no reconstruction at all (large $r$) without
suffering too much loss. Hence this loss function achieves the goal that
only some of the input dimensions are
\textit{selectively reconstructed}, instead of all of them. The exact dimensions
that are reconstructed are chosen automatically by the learning procedure
itself.}

Finally, to make the x-features in the explanation space more orthogonal
to each other and avoid several x-features explaining the same or similar
concepts, an orthogonality loss is added to the optimization. Here we utilize
the pull-away term (PT) \cite{pullaway17} that has been successfully applied
in generative adversarial networks:
%
\begin{align}
L_{PT} = \frac{1}{n(n-1)}\sum _{l=1}^{n}\sum _{l'\neq l}{\Big (}
\frac{\mathbf{E}_{l} ^{T} {\mathbf{E}}_{l'}}{\left \|  {\mathbf{E}}_{l}\right \|  \left \|  {\mathbf{E}}_{l'}\right \|  }
{\Big )}^{2}
\label{eq:pt}
\end{align}%
where $n$ is the number of x-features,
$\mathbf{E}_{l} = \mathbf{E}_{\bm{\theta }}(\mathbf{Z})_{l}$ represents the vector
for the $l$-th x-feature over the training set. We define the
final optimization problem as:
%
\begin{equation}
\label{eq:totalLoss}
\centering
L_{SRAE} = L_{FS} + L_{SR} + L_{PT}
\end{equation}
where SRAE (Sparse Reconstruction Autoencoder) is our proposed model which
handles the faithfulness, locality, and orthogonality objectives. SRAE
is a neural network hence can be seamlessly combined with the prediction
DNN, making the following visualization process (introduced in Sec.~\ref{sec:visualization})
simple.


In practice, we utilize L2-loss for $L_{FS}$ and the log penalty for $L_{SR}$, and rewrite the
optimization of the SRAE as follows:
%
\begin{align}
\label{eq:saeff}
\min _{\bm{\theta },\bm{\widetilde{\theta }},\mathbf{v}}&\frac{1}{M}\sum _{i=1}^{M}{
\Big \|}{\mathbf{v}}^{\top }{\mathbf{E}}_{\bm{\theta }}(\mathbf{Z}^{(i)})-\hat{\mathbf{y}}^{(i)}{
\Big \|}^{2} + \beta \cdot \frac{1}{S_{z}} \sum _{k=1}^{S_{z}}
\text{log}(1+q\cdot\frac{1}{M}\sum _{i=1}^{M} {\Big \|}{\mathbf{E}}^{-1}_{
\bm{\widetilde{\theta }}}\Big (\mathbf{E}_{\bm{\theta }}(\mathbf{Z}^{(i)})
\Big )_{k}-{Z}^{(i)}_{k}{\Big \|}^{2})
\nonumber
\\
&+\eta \cdot \frac{1}{n(n-1)}\sum _{l=1}^{n}\sum _{l'\neq l}{\Big (}
\frac{\mathbf{E}_{l} ^{T} {\mathbf{E}}_{l'}}{\left \|  {\mathbf{E}}_{l}\right \|  \left \|  {\mathbf{E}}_{l'}\right \|  }
{\Big )}^{2}
\end{align}%
where the 3 terms are faithfulness, sparse reconstruction, and orthogonality,
respectively. 
$\beta , \eta $ are weights for each respective term. The prediction result
$\mathbf{v}^{\top }{\mathbf{E}}_{\bm{\theta }}(\mathbf{Z}^{(i)})$ of SRAE is denoted later
as ${\bar{\mathbf{y}}}^{(i)}$.

Fig.~\ref{fig:XDLOrth}(a) shows the XNN structure. $\bm{\theta }$ performs the embedding
from the input layer $\mathbf{Z}$ to the explanation space $\mathbf{E}$. From
$\mathbf{E}$, $\tilde{\bm{\theta}}$ performs reconstruction, and $\mathbf{v}$ maps to the
prediction layer $\hat{\mathbf{y}}$ to be faithful to the DNN prediction. The pull-away
term ensures the orthogonality in $\mathbf{E}$. With a trained XNN we would obtain
an explanation embedding $\mathbf{E}_{\bm{\theta }}(\mathbf{Z})$ and a linear
predictor $\mathbf{v}^{\top }{\mathbf{E}}$ so that the output of the network can be
explained with a weighted sum of x-features
$\mathbf{E}_{\bm{\theta }}(\mathbf{Z})$ with the weight $\mathbf{v}$ indicating the
contribution of each individual x-feature to the final prediction. In conjunction
with the visualization paradigm in the next subsection, this facilitates
a better understanding of the black-box DNN model.


\subsection{Visualizing the Explanation Space}
\label{sec:visualization}
The goal in the visualization of low-dimensional explanation features is 
to bridge the communication gap between human and machine, enable humans to name concepts learned by the explanation network, and be able to construct sentences with those named concepts in the future. In the scope of this paper, we only focus on visualizing the concepts.

We utilized ExcitationBP \cite{zhang16excitationBP} to compute the contrastive marginal winning probability (c-MWP) from each neuron in the x-layer to the pixels in the original image, then generate the heatmaps using c-MWP normalized on each neuron for each image. 
The reason for choosing this algorithm over competitors such as Grad-CAM \cite{selvaraju2016gradcam} or RISE \cite{Petsiuk2018rise} is that this algorithm gives more detailed predictions that are not blurry or hard for humans to interpret.

Figure \ref{fig:IntrEx} shows some examples of the visualization results of these concepts learned by the proposed explanation network. The number above each heatmap represents the contribution of this x-feature to the final prediction.
For instance, the left example in Figure \ref{fig:IntrEx} shows that there are two key features to predict {\em European goldfinch}: x-feature \#$1$ ({contribution: $63.1030$}) and x-feature \#$2$ ({contribution: $15.1854$}) where x-feature \#$1$ is much more important than x-feature \#$2$ based on their contribution to the final prediction. 
Note that our approach generates the visualizations for the human to deduct those features, without requiring any textual annotation to train. For the case of Figure \ref{fig:IntrEx}, humans can study the visualizations and then name x-features \#$1$ and \#$2$ as {\em golden feather} and {\em red forehead}, respectively.

\section{Related Work}

The explanation for high accuracy but black-box models has become a significant need in many real applications. A large amount of approaches have been proposed in the past few years in medical domain \cite{Caruana15,letham2015,ustun2016}, Natural Language Processing \cite{kulesza2015,Ribeiro2018AnchorsHM,Sarah19Attention}, computer vision, etc.
In computer vision, approaches have been introduced to explain the predictions either by associating the images with captions/descriptions \cite{kiros2014multimodal,kong2014you,lin2014visual,Karpathy_2015,hendricks2016generating}, visualizing individual convolutional filters in the network~\cite{zeiler2014v,bau2017network} or heatmaps that indicate important regions in the original images \cite{simonyan2014deep,cao2015look,zhou2016learning,zhang16excitationBP,selvaraju2016gradcam,Bach2015OnPE,kindermans2017,GradCamPlus17,Shrikumar2017LearningIF}. \cite{park2016attentive} and \cite{selvaraju2016gradcam} propose to explain via visual question answering which utilized both natural language descriptions and heatmaps.
\cite{ribeiro2016should} proposes an explanation technique that tries to explain a single prediction of general models and select several representative predictions to provide a global view of the model.
\cite{ScottNIPS2017} and \cite{ElenbergNIPS17} propose a unified approach and a streaming algorithm, respectively, to interpreting model predictions.

Image captioning approaches~\cite{kiros2014multimodal,kong2014you,lin2014visual,Karpathy_2015,hendricks2016generating} need to be trained on human-generated sentences, hence they would not work in any domain where human is not an expert in. Our approach does not require any natural language descriptions. 
Visualizing individual neurons/filters were important for human intuition about CNNs \cite{zhou2015object,zeiler2014v,jain2016structural}. \cite{bau2017network} went to great lengths in visualizing thousands of neurons  and asking humans to name each of them. However, it is difficult for such efforts to provide a concise yet complete representation. \cite{agrawal2014analyzing} analyzed the number of filters required to generate good performance on the PASCAL VOC dataset and the conclusion is that each class would need at least dozens of filters. \cite{che2016interpretable} learns a decision tree on top of a deep network as an attempt for an explainable model. And similar to us, \cite{zhou2018interpretable} proposed a framework to decompose the activation feature vector of a network into several semantic concepts. These 2 approaches train on an existing vocabulary of attributes, which are not able to achieve very high faithfulness.
We adopt the heatmap approach in~\cite{zhang16excitationBP}, but visualize explanation features instead of directly visualizing classification results. With this approach, we can generate high-level concepts that are broadly applicable to multiple images in the same category.

There has been a focus on detecting parts using deep neural networks without part annotations, usually in fine-grained classification \cite{simon2015neural,xiao2015application,gkioxari2015actions,zhang2016spda,zhang2016weakly,zhang2016picking}. 
Our focus is different in that we focus on explaining a trained deep model instead of trying to enhance it, and explanations may not necessarily be parts that are usually expressed in terms of bounding boxes as in those approaches.
\cite{gonzalez2016semantic} conducted comprehensive experiments on whether semantic parts naturally emerge from convolutional filters. They explored combinations of filters using a genetic algorithm but only combine an average of $5$ filters, hence did not have the dramatic dimensionality reduction effect as in our work. 
\cite{zhao2016diversified} train a hybrid CNN-LSTM model featuring diversified attention models jointly and generate diverse attention maps similar to ours in the middle of the network, but it cannot be utilized to explain an already-trained DNN because of the joint training that is needed.

Model compression for deep learning was proposed in~\cite{ba2014really}, where a shallow model is used to mimic the output of a deep network. Most model compression work  were used for speeding up inference~\cite{chen2015compressing,rastegari2016xnor} instead of explanation. 
\cite{OramasArxiv17,WangRoutingCVPR} are sparse feature selection methods. Both use L1 regularization (in \cite{OramasArxiv17}, from different layers) to select nodes from the DNN. XNN is instead feature extraction. Both of \cite{OramasArxiv17,WangRoutingCVPR} kept up to 10-15$\%$ of the original features (about 28 (MNIST) to 248 (ImageNet)). It’s hard to visualize these many features to humans for the user study we did. Hence, we believe XNN is significantly better than these 2 in conciseness.
\cite{ZhangICNN17} and TCAV \cite{Kim2018InterpretabilityBF} quantify the degree to which a user-defined concept is important to a classification result. Different from them, our approach does not rely on predefined concepts.

\cite{OscarCase18,DavidSelf18,Chen2019ThisLL} propose self-explaining networks which are based on prototypes, but they need to train new networks, while XNN explains an already-trained network. 
After the publication of an earlier version of XNN \cite{QiNIPSLearning17}, \cite{TanDis18} propose an approach for explaining complex models, but is more suitable for neural nets trained on tabular data. It cannot be applied on complex image datasets. 

In the unsupervised learning domain, \cite{InfoGAN,Kumar18} train networks to learn disentangled representations for simpler, structured images. Most of these approaches only work for simpler images such as rendered models and handwritten digits, while our approach focuses on natural images, which has a more complex and challenging distribution~\cite{Ruderman94natural}. 
\cite{Ronny19} learns high-level features by optimization on every single image, which is more like the perturbation based heatmap approaches. However, our method learns high-level concepts for a general category. 
\cite{AndyBayesian18} focuses on explaining Bayesian network classifiers, which compiles latent-tree Bayesian network classifiers into decision functions; \cite{AlexeyAbduction19} focuses on exploiting abductive reasoning for computing explanations of ML models with formal guarantees, e.g. cardinality-minimal or subset-minimal explanations. None of them generate concise and high-level concepts like XNN though. 
\cite{molnar2019} summarizes the concepts and approaches in the area of interpretable Machine Learning.

A lot of previous studies \cite{Lipton2018TheMO,DoshiVelez2017TowardsAR,Rudin2018StopEB,Riedl2019HumanCenteredAI,Ross2017RightFT} try to refine the discourse on interpretability, define different concepts on it, and describe its characteristics and properties.
\cite{Sokol2020ExplainabilityFS} introduces a taxonomy and a set of descriptors that can be used to characterise and systematically assess explainable systems along five key dimensions: functional, operational, usability, safety, and validation. The taxonomy includes more than 30 requirements hence it is impossible for us to list them all. Among the functional requirements, XNN is a post-hoc explanation of supervised learning algorithms, focuses on explaining numerical models, applicable to both black-box classification and regression models, and is model agnostic. Operationally, it looks for associations and deliver visualizations to the user in a static manner, requires a transparent model, and does not reduce model performance. In terms of usability, XNN maintains soundness (faithfulness), is complete (demonstrated in experiments) and parsimonious. One of the drawbacks is that it might be hard for users to interpret since its concepts do not always correspond to human concepts.


\section{Experiments}
\subsection{Human Evaluation}
To investigate the effectiveness of the proposed XNN with SRAE, we designed a user study where participants were asked to cluster images that are normally difficult to discern by untrained humans{\red , referred to as \textit{main human study}}. The goal was to inspect whether the XNN can provide interpretable visualizations to non-experts so they can differentiate between unlabeled samples about which they do not have prior knowledge about. We compare \textbf{XNN} against two baselines, one without any visualization (\textbf{NoVis}), and one with a single heatmap derived from the original output $\hat{\bf y}$ (\textbf{1-Heatmap (ExcitationBP)}). For fairness, the same heatmap approach, ExcitationBP~\cite{zhang16excitationBP} is used throughout the study. This approach is chosen since it obtains detailed heatmaps, rather than some other performant approaches which are more blurry~\cite{selvaraju2016gradcam}. {\red In addition to the main study, we held a \textit{complementary human study} to compare the previously obtained results from the above approaches against another single heatmap baseline using DeepLIFT~\cite{Shrikumar2017LearningIF} method, referred to as \textbf{1-Heatmap (DeepLIFT)}. Note, the study design and procedure were exactly as in the main study but with different participants and a different visualization method in the 1-Heatmap stage.
}

\begin{figure*}[t]
\vspace{-0.18in}
\begin{center}
\subfigure[] { \label{fig:snapNOVIS}
\includegraphics[width=0.44\linewidth]{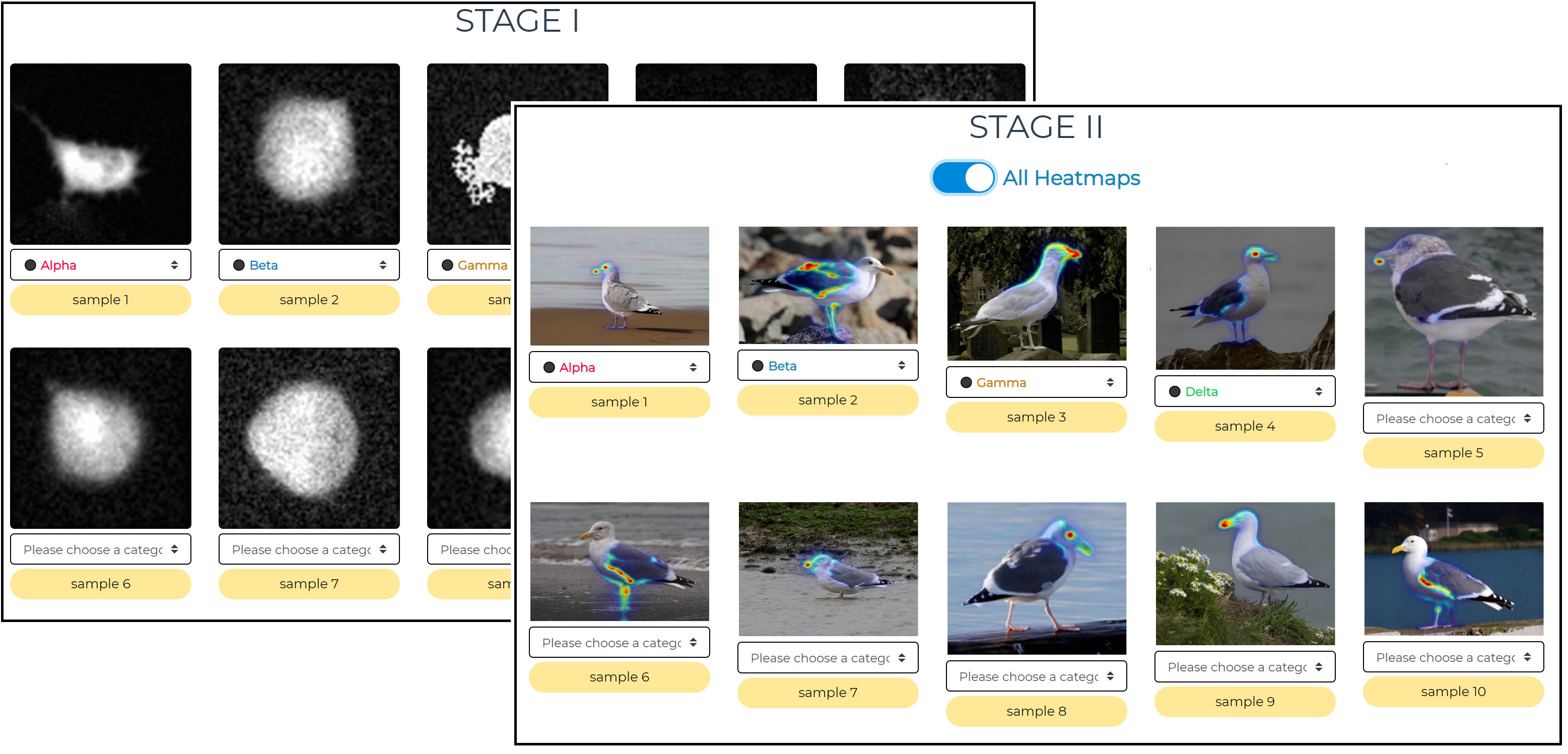}
}
\subfigure[] { \label{fig:snapEXBP}
\includegraphics[width=0.52\linewidth]{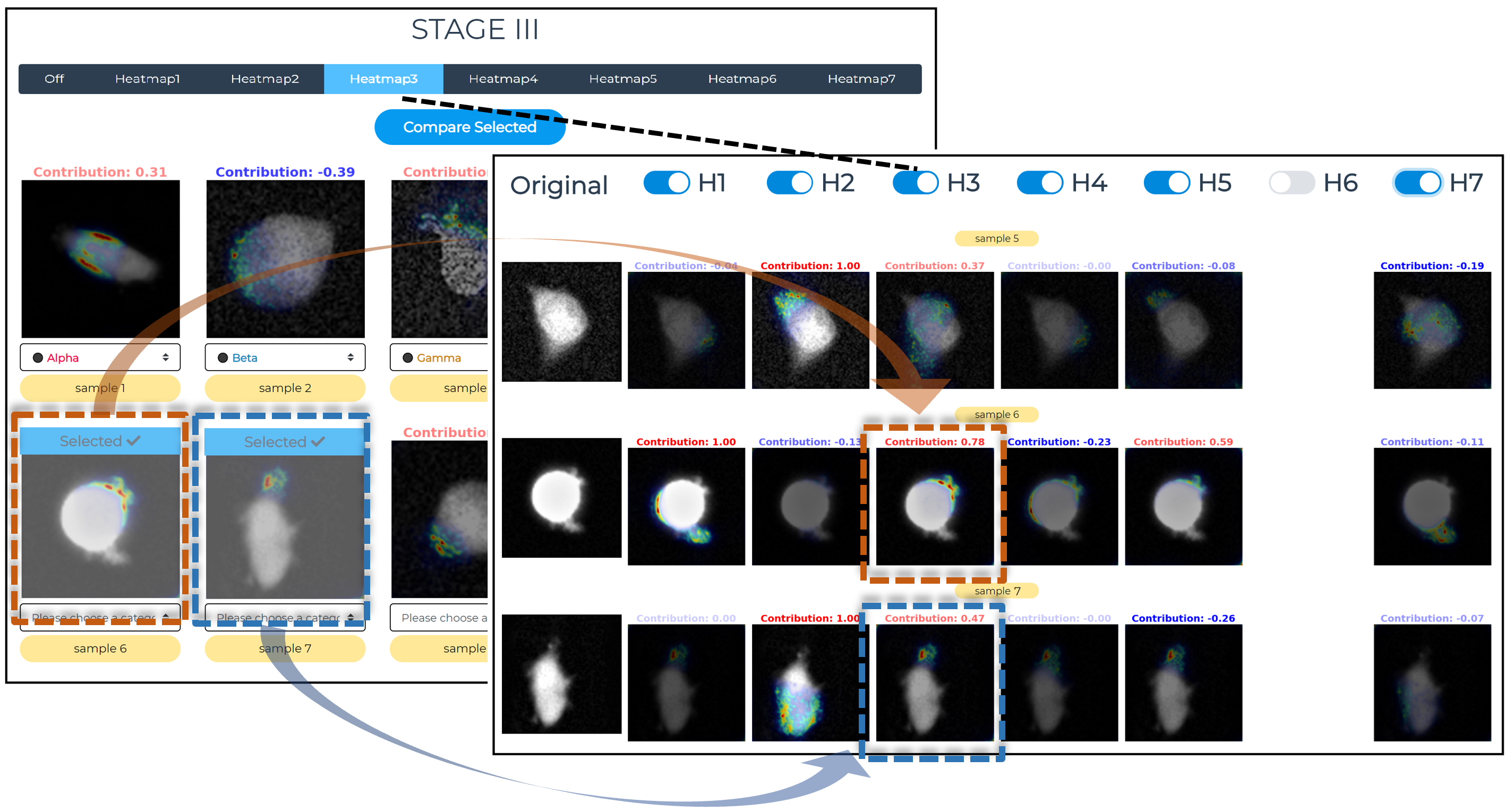}
}
\end{center}
\vspace{-0.25in}
   \caption{\small Snapshots from different stages of our designed human interface. (a) NoVis stage with CEL images (back) and 1-Heatmap stage with CUB images (front). In the 1-Heatmap stage, users could either hover over the sample images to see the ExcitationBP generated heatmaps or toggle on/off heatmaps for all images at once using a button in the interface (b) the XNN stage with another batch of CEL (back) images where the users could toggle through all the visualizations from x-features 1-7. In this snapshot, {\em Heatmap 3} is toggled on. The value on top of each heatmap indicates the normalized contribution of that x-feature to the final prediction (positive values in red and negative values in blue). Also, users could select images by clicking on them (the ones with blue banner on top) and compare their heatmaps from all x-features in one place (front). Here, {\em Heatmaps (H) 1-5\&7} are toggled on and {\em Heatmap 6} is toggled off.}
\vskip -0.15in
\label{fig:snap}
\end{figure*}


We selected 4 categories of image samples from 2 different datasets where the distinction among the categories is not obvious to non-experts: first, 4 categories of visually alike seagulls (California Gull, Herring Gull, Slaty-backed Gull, and Western Gull) were selected from the CUB-200-2011 dataset \cite{WahCUB200}, referred to as ``CUB'' in the rest of the paper. 
Second, 
we also selected 4 categories of breast cancer cells (Actinedge, Hemispherebleb, Lamellipodia, and Smalbleb) from microscopic images~\cite{eddy2018morphodynamics}, referred to as ``CEL''. 
In human study, each XNN is multi-class classifier with 4 classes, where we set the number of the x-features as $7$. 
For each dataset, 200 images were selected where each category contained 50 images. 


\begin{figure}[t]
\begin{center}
\subfigure[] { \label{fig:resCEL}
\includegraphics[width=0.41\linewidth]{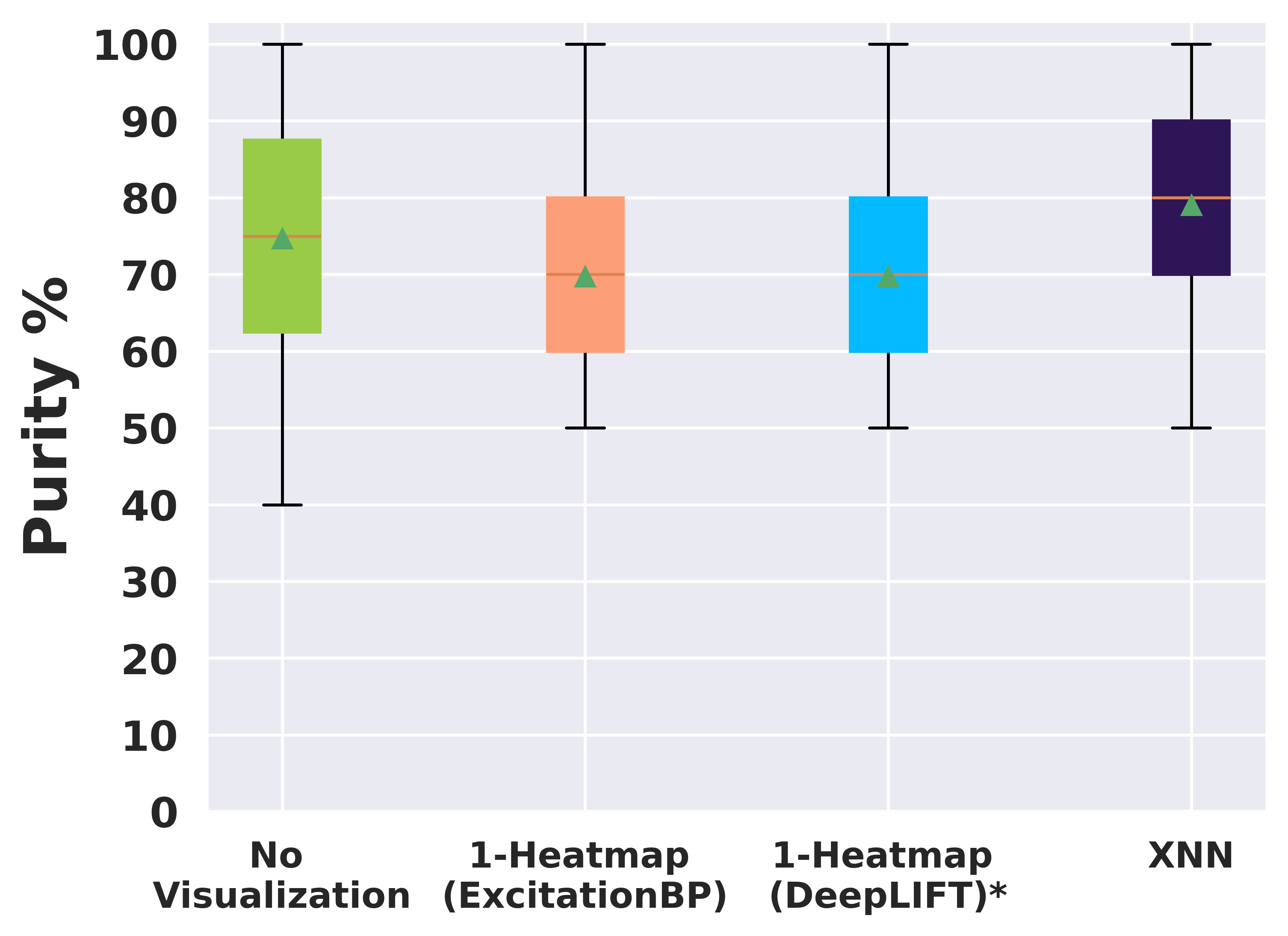}
}
\subfigure[] { \label{fig:resCUB}
\includegraphics[width=0.41\linewidth]{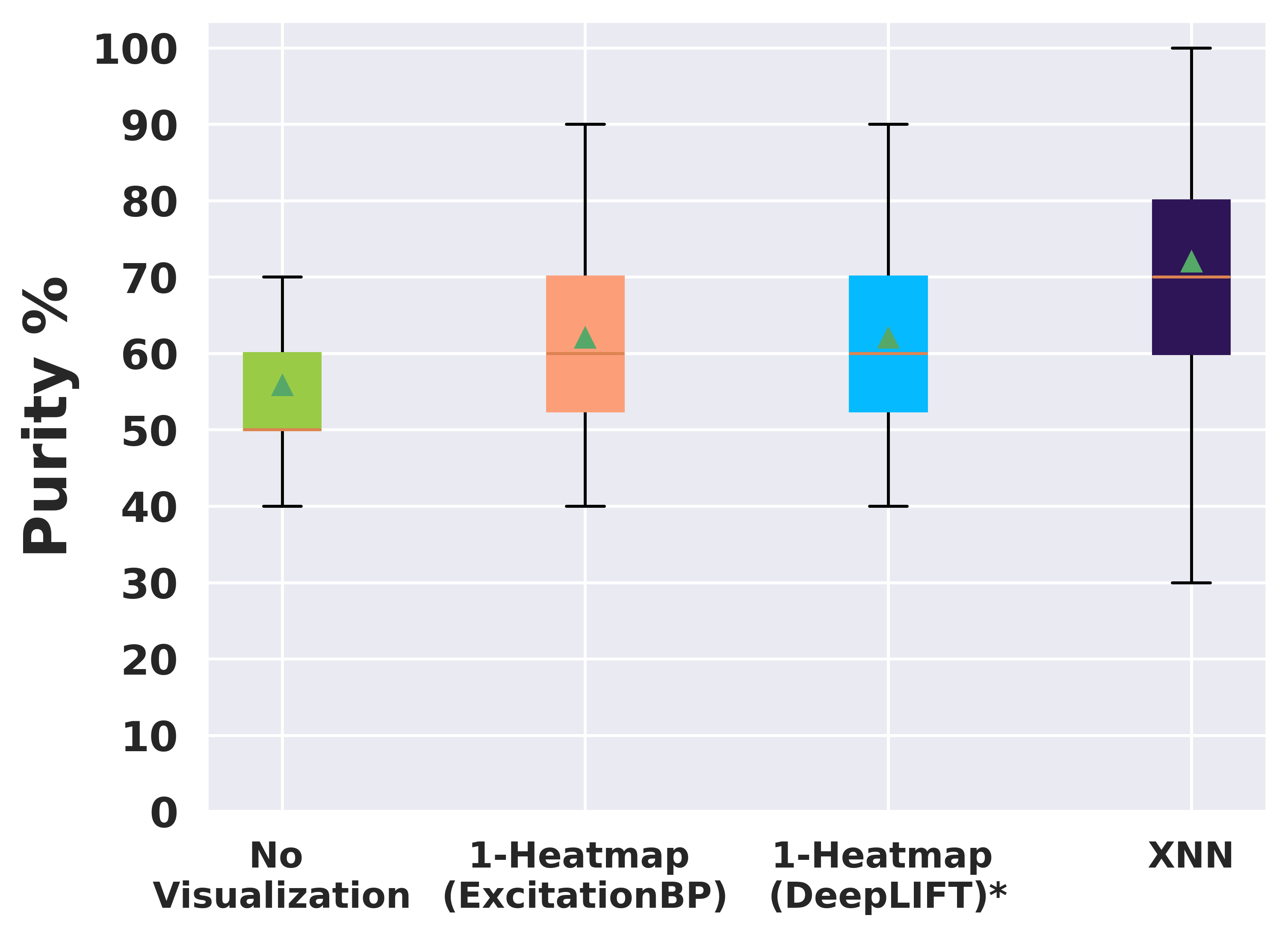}
}
\end{center}
\vspace{-0.25in}
   \caption{\small Boxplot of clustering purity with respect to the original CNN prediction obtained from the human evaluation for NoVis, 1-Heatmap (ExcitationBP), 1-Heatmap (DeepLIFT)*, and the XNN on the (a) CEL dataset and (b) CUB dataset. {\red (*1-Heatmap (DeepLIFT) results are from the complementary human study rather than the main human study.)}}
\vspace{-0.15in}
\label{fig:res}
\end{figure}


The evaluation interface comprises of 3 main stages: {\bf NoVis} (No Visualization) where the participants were not provided with any visual explanation and they were asked to cluster images only based on their own beliefs of the differences and/or similarities among the samples. 
\textbf{1-Heatmap} is the baseline where the participants were provided with heatmaps derived from the original output $\hat{\bf y}$ using ExcitationBP {\red or DeepLIFT} to assist them on clustering. The participants were told that the AI agent utilizes these heatmaps to distinguish one sample from another. {\bf XNN} refers to our method where for each sample image, users were provided with different heatmaps generated by ExcitationBP corresponding to each x-feature to help them in decision making. They were also told that a contribution of all the different heatmaps was used by the AI agent when trying to identify sample images. Snapshots of our user interface at the aforementioned stages can be found in Figure \ref{fig:snap}. {\red In the complementary human study the participants were needed to first pass the NoVis stage before continuing to the 1-Heatmap stage where they were provided with heatmaps from the DeepLIFT. This choice was to keep the human study conditions as close as possible to the main one.}

For each stage, users were shown 2 batches containing 10 sample images, one from the CUB dataset and the other one from the CEL dataset. 
To reduce randomicity, 5 sets of 10 images each were drawn randomly at first, and the batch showed to the users were randomly selected from these 5 sets.
In addition, the order whether XNN or 1-Heatmap was shown first is random, the order whether CEL or CUB is shown first is also random, but the dataset would always be alternated. For instance, one possible ordering for the stages and the samples shown {\red in the main study} can be as following: {\em (NoVis-CEL, NoVis-CUB, XNN-CEL, XNN-CUB, 1-Heatmap-CEL, 1-Heatmap-CUB).} 
Hence, the users had to alternate between the 2 datasets to reduce memorizations of the previous visualizations they have seen on a dataset. 
For both datasets, the visualizations shown in the XNN stage were generated from the 7 x-features that were obtained from embedding the high-dimensional feature space using our proposed model formulated in section \ref{sec:mFormulation}.


{\red Results from the main human study (from 30 participants) as well as the results from the \textbf{1-Heatmap (DeepLIFT)} obtained in the complementary human study (from 14 participants) are shown in Figure \ref{fig:res}(a-b)}. We measured purity of user clusterings with respect to the original predictions of the network in each stage. For the CUB dataset, \textbf{XNN} achieved purity of 72\% while \textbf{NoVis}, \textbf{1-Heatmap {\red (ExcitationBP)}}{\red, and \textbf{1-Heatmap (DeepLIFT)}} achieved 56\%, 62\%, and 57.8\% respectively. 
For the CEL dataset, mean purity of 75\%, 69.6\%, 69.2\%, and 79\% was obtained for \textbf{NoVis}, \textbf{1-Heatmap {\red (ExcitationBP)}}, {\red \textbf{1-Heatmap (DeepLIFT)}} and \textbf{XNN}, respectively (Fig. \ref{fig:resCEL}). The difference of the results between the \textbf{XNN} and \textbf{1-Heatmap} {\red approaches} was statistically significant for both datasets, which means XNN helps human understand more on these difficult classification tasks than just seeing one heatmap.
%
The results show that humans are moderately successful in classifying the CEL dataset even if no visualization is shown, but struggles significantly to distinguish the 4 types of seagulls in the CUB dataset. XNN significantly improved their performance in CUB by pointing out the exact features that are salient to separate the birds which are otherwise very similar. In CEL, XNN notched some improvements over \textbf{NoVis}, but not enough to be statistically significant, whereas \textbf{1-Heatmap {\red (ExcitationBP)}} {\red and \textbf{1-Heatmap (DeepLIFT)}} seem to have even hurt human performance a bit.



\begin{figure*}[t]
\begin{center}
\subfigure[] { \label{fig:confCEL}
\includegraphics[width=0.32\linewidth]{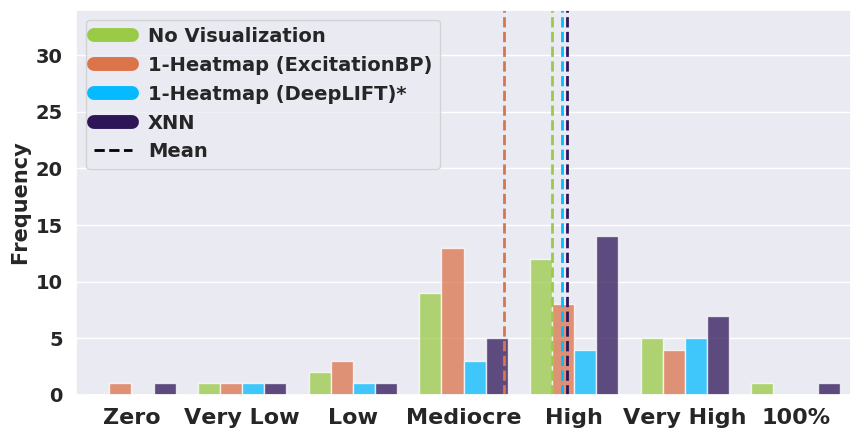} 
}
\subfigure[] { \label{fig:confCUB}
\includegraphics[width=0.32\linewidth]{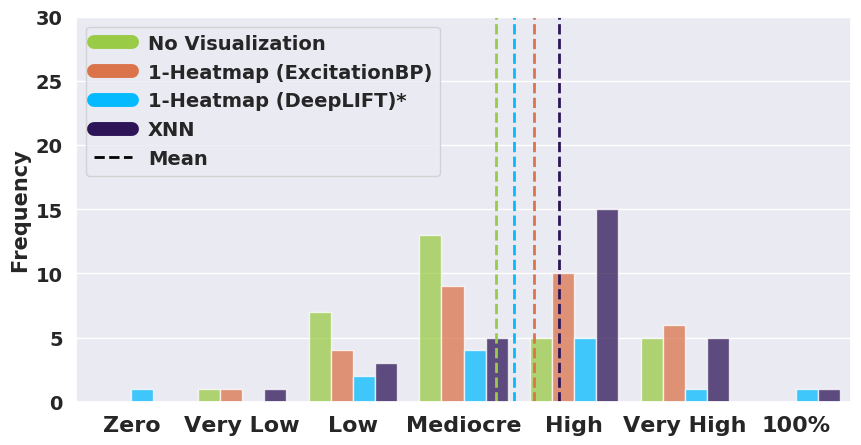}
}
\subfigure[] { \label{fig:perf}
\includegraphics[width=0.3\linewidth]{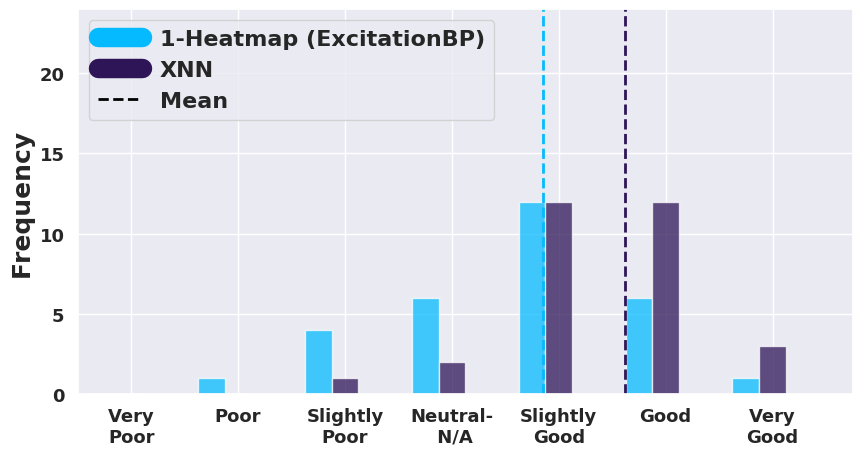} 
}
\end{center}
\vspace{-0.25in}
   \caption{\small Participants' confidence level in different stages that their answers would correctly group all the sample images for the (a) CEL dataset (b) CUB dataset. 
(c) Participants' answers to the question that how do they evaluate the performance of the 2 different AI agents helping them on grouping images at the end of the main human study, 1-Heatmap {\red (Excitation BP)} as the baseline (in {\red blue}) and XNN as our approach (in {\red purple}).  {\red (*1-Heatmap (DeepLIFT) results are from the complementary human study rather than the main human study.)}}
\vspace{-0.15in}
\label{fig:conf}
\end{figure*}

\textbf{Confidence Levels}: At the end of each stage of the human studies, the participants were also asked the question "{\em What is your confidence level that the answers you picked will group all the samples correctly}?" with Likert scale answers: {\em ('Zero', 'Very Low', 'Low', 'Mediocre', 'High', 'Very High', '100\%')}. The obtained results are illustrated in Figure \ref{fig:conf}. It can be seen that the users' confidence levels in the validity of their answers were in accordance with the actual clustering results that were obtained. For the CEL dataset, participants' confidence in the XNN stage was marginally higher than the 1-Heatmap stage (p-value: 0.083). However, for the CUB dataset, the difference of the visualization approaches is not noticeable yet both of them seem to elevate the confidence level of the participants compared to no visualization approach.


\textbf{Performance Comparison}: At the end of the main study, we handed out printed questions to the participants and asked them to rate the performance of the 2 different AI agents (visualization methods) that were assisting them during the study, the EcxitaitonBP on the class (baseline) and x-features (our approach) in distinguishing images from each other. The users could rate the performance in a Likert scale with the options: {\em ('Very Good', 'Good', 'Slightly Good', 'Neutral or N/A', 'Slightly Poor', 'Poor', 'Very Poor')}. The obtained results are depicted in Figure \ref{fig:perf}. This shows that humans tend to significantly prefer the performance of our approach over the baselines (p-value: 0.0049).



\subsection{Quantitative Evaluation Metrics}
We believe that evaluating explanations objectively without a human study is also important because simple parameter variations can easily generate thousands of different explanations, vastly outpacing the speed of human studies. In this paper, we make an attempt to define some quantitative metrics without human evaluation. 
We utilize the CUB-200-2011 dataset \cite{WahCUB200} and Places365 dataset \cite{zhou2017places} in our quantitative experiments. 
The former dataset is a fine-grained classification task with $200$ categories of birds. From a network trained on all the $200$ categories, we attempt to explain the logits (before the softmax layer) of each category separately with one XNN per category. This dataset is chosen because in addition to category labels and bounding boxes surrounding each object, it also has part labels denoted as one pixel per part for each object as additional ground truth. 
One can argue that the majority of bird classifications are based on specific, discriminative parts of the bird, which can be confirmed from  encyclopedias and expert annotations~\cite{reed2016learning}. 
The latter dataset (Places) is a task for scene understanding and recognition, where we selected $10$ room categories from the ADE20K dataset \cite{zhou2017scene}. 
This dataset is chosen because, besides the scene labels, there also exist segmentation labels as additional ground truth for some images in some categories.

\begin{figure}[t]
\begin{center}
\includegraphics[width=0.65\linewidth]{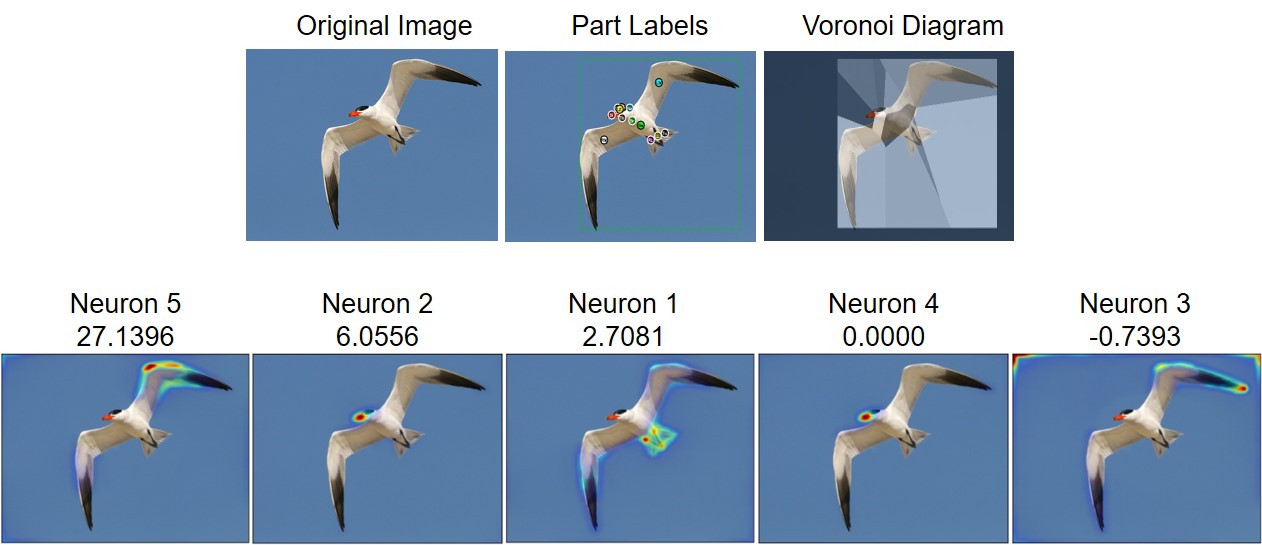}
\end{center}
\vspace{-0.25in}
   \caption{\small A simple example is generated by the explanation module. The first line shows the original image, the part labels of the image in the ground truth, and the Voronoi diagram of the image; the second line shows the visualization results for the 5 neurons in the x-layer sorted by the weights (${v}_i{E}_i, i=1,2,\ldots,5$) for the final prediction.}
\vspace{-0.15in}
\label{fig:example}
\end{figure}

\begin{figure}[t]
\centering
\subfigure[] { \label{fig:Scora}
\includegraphics[width=0.47\linewidth]{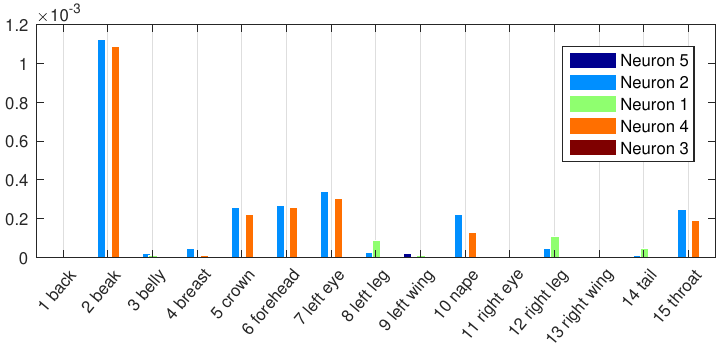}
}
\subfigure[] { \label{fig:Scorb}
\includegraphics[width=0.47\linewidth]{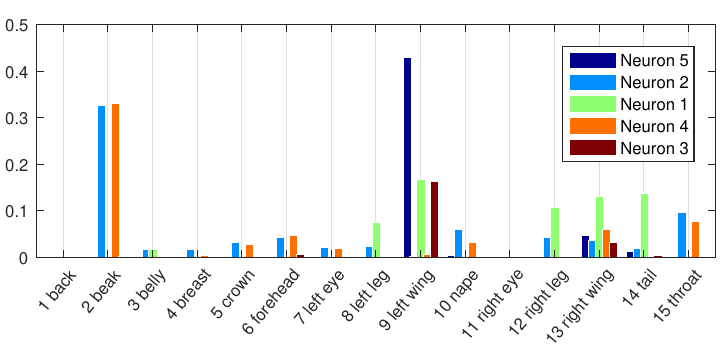}
}
\vskip -0.1in
\caption{\small (a) Pixel-level probability $S_{i_p,j_p}^{n,m}$; (b) Voronoi-based probability $S_{p}^{n,m}$ for the example image in Figure  \ref{fig:example}.}
\vskip -0.1in
\end{figure}


For the CUB-200-2011 dataset, the fine-tuned VGG19 model
\cite{Simon15} is used as the prediction DNN to be explained. For the Places365
dataset, the fine-tuned VGG16 model \cite{zhou2017places} is used as the
prediction DNN to be explained. The explanation network is a 3-middle-layer
SRAE with $800, 100, n$ hidden units in each layer, respectively, where
$n$ represents the number of x-features. 
In quantitative evaluations, each XNN is a binary one-vs-all classifier which is used to interpret the results of only one category.
For CUB, we trained an explanation
network on each of the $200$ bird categories. For each category, we utilized
$50$ positive examples and $8,000$ negative examples as the training data;
the remaining positive examples ($8-10$) and $2,000$ negative examples
as the testing data. In the training process, we enhance the weights of
the positive examples to avoid imbalance. $n$ is set to $5$, as our experiments
showed that more x-features do not improve performance and create x-features
that have $0$ weight in $v_{i} E_{i}$, indicating that the one-vs-all
classifier of one bird may not depend on many high-level visual features.
For Places, we trained an explanation network on each of the $10$ scene
categories which are different kinds of rooms and have enough images with
object labels in ADE20K to evaluate, including \emph{bathroom, bedroom,
conference room, dining room, home office, hotel room, kitchen, living
room, office, and waiting room}. For each category, we utilized
$4,000$ positive examples and $20,000$ negative examples as the training
data; about $1,000$ positive examples and $4,000$ negative examples as
the testing data. $n$ is also set to $5$ for Places. 
We compared the proposed SRAE with a fully-connected neural network (NN) (no reconstruction), a conventional stacked autoencoder {with faithfulness loss and traditional reconstruction loss} (SAE) (no sparsity term), {a classic autoencoder with only traditional reconstruction loss and without faithfulness loss (CAE)}, a feature selection model (Lasso) on $\textbf{Z}$, as well as directly performing ExcitationBP on the classification output $\hat{\mathbf{y}}$ (1-Heatmap). 
The baseline neural network methods (NN and SAE) can also perform a faithful dimensionality reduction and are the most closely related to our approach. {Lasso represents a feature selection approach which selects several most useful dimensions directly from $\textbf{Z}$ and tries to mimic the network decision as a linear combination of these features.} All the learning-based approaches (SRAE, NN, SAE, {CAE,} and Lasso) were tuned to the optimal parameters by cross-validation on the training set.

Given image $I_m$, for each neuron $n$ in the x-layer and each pixel $({i,j})$ in $I_m$, we denote $\small{S_{i,j}^{n,m} \triangleq P(\text{Pixel}^m_{i,j}|\text{Neuron}_n) =\frac{C_{i,j}^{n,m}}{\sum_{(i,j)\in I}C_{i,j}^{n,m}}}$, where $C_{i,j}^{n,m}$ is the c-MWP generated by ExcitationBP for pixel $({i,j})$ in $I_m$ with neuron $n$ in x-layer, $({i,j})$ is the coordinate of the pixel. 
For the CUB dataset, the given part label ($p=1,...,15$) of each image is just one pixel in the middle of the part, and there is no extra information about the shape and the size of the part regions.
For the $p$-th part label of image $I_m$, we denote $(i_p, j_p)$ as its pixel location.
The {\bf pixel level probability} is defined as $S_{i_p,j_p}^{n,m}=P(\text{Pixel}^m_{i_p,j_p}|\text{Neuron}_n)$.
Figure \ref{fig:Scora} shows the probability $S_{i_p,j_p}^{n,m}$ for each neuron $(n=1,\ldots,5)$ at the pixel locations of the part labels $(p=1,\ldots,15)$ for the example image shown in Figure \ref{fig:example}. 
From Figure \ref{fig:Scora} we observe that the probability $S_{i_p,j_p}^{n,m}$ is reasonable when capturing small parts like {\em beak}, but is not on larger parts like {\em wing}, for the part label is just one pixel in the middle of the {\em wing}, while the x-features mainly focus on the edges (Fig.~\ref{fig:example} shows a simple example).


Thus, we utilize the Voronoi diagram~(Fig.~\ref{fig:example}) \cite{Voronoi} to partition the bounding box into $15$ regions in which the nearest neighbor part annotation in each region would be the same,
and then compute the  {\bf Voronoi-based probability} $S_{p}^{n,m} \triangleq P(\text{Part}^m_p|\text{Neuron}_n)=\sum_{(i,j)\in I_m}P(\text{Part}^m_p|\text{Pixel}^m_{i,j}) P(\text{Pixel}^m_{i,j}|\text{Neuron}_n)$.
In mathematics, a Voronoi diagram is a partition of a plane into regions close to each of a given set of objects. 
The Voronoi diagram is used instead of segmentation because we do not have segmentation ground truth and do not wish to include additional errors from an arbitrary segmentation algorithm. Besides some of the heatmap activations fall slightly outside the object and we still want to capture those.
However, the larger parts such as {\em wing} and {\em tail} always obtain much higher scores than the smaller parts such as {\em beak} and {\em eye} do; and there are also many background pixels far from the center contained in the Voronoi diagram.
To solve these issues, we introduce the inverse distance as a factor when computing the {\bf Voronoi-based probability} $S_p^{n,m}$ in Algorithm 1 in the Appendix, trying to keep the balance between the large part region and the small part region.


Figure \ref{fig:Scorb} shows the probability $S_{p}^{n,m}$ for each neuron and each part label for the same example image in Figure \ref{fig:example}.
From Figure \ref{fig:Scorb} one can also see evidence that 
the probabilities on {\em wing} and {\em tail} of some neurons are higher, indicating the metric based on the Voronoi diagram enhances the evaluation of these larger parts. 
For all the c-MWP outside of the ground truth bounding box, we introduce a $16$-th part called \textit{context}, which indicates that the x-feature is using the context to classify rather than the object features. 
For the Places dataset, since we have the exact object regions for different object labels of each image, we compute the probability $S_{p}^{n,m} \triangleq P(\text{Object}^m_p|\text{Neuron}_n)=\sum_{(i,j)\in \text{Object}^m_p} P(\text{Pixel}^m_{i,j}|\text{Neuron}_n)$. 
For each x-feature $n$ we have a histogram ${\bf S}_n$ whose element is ${\bar S}_{p}^n= \frac{1}{M}\sum_m S_{p}^{n,m}$.

\begin{table}
    \centering
    \caption{\small The average faithfulness, orthogonality, and locality of different approaches over all the $200$ categories of the CUB dataset. The column ${\bf Z}$ represents the average locality computed over all the dimensions of ${\bf Z}$, the $4,096$-dimensional first fully-connected layer of the deep network. This is obtained by separately running ExcitationBP on each dimension of ${\bf Z}$ and evaluating the resulting heatmaps.  1-Heatmap refers to the heatmap from $\hat{y}$.}
        \vspace{-.07in}
{\small
\begin{tabular}{||@{  }c@{  }|@{  }r@{  }|@{  }r@{  }|@{  }r@{  }|@{  }r@{  }|@{  }r@{  }|@{  }r@{  }|@{  }r@{  }|@{  }r@{  }||}
\hline
  \multicolumn{2}{||l|}{Method} &SRAE     & NN       & SAE &Lasso & {CAE}  & $\bf Z$ & 1-Heatmap \\\hline\hline
$F_{reg}$ & Train & 0.0812 & 0.0696 & 0.0972 & 3.5785 &{4.1513} & --- & --- \\
& Test & 0.1659 & 0.1304 & 0.1981 & 3.7928 &{4.0021} & --- & --- \\\hline
$F_{cls}$ & Train & 99.99\% & 100.0\% & 99.99\% & 73.14\% &{65.34\%} & --- & --- \\
& Test & 99.99\% & 100.0\% & 99.98\% & 71.53\% &{69.28\%} & --- & --- \\\hline\hline
O1 & & {0.6554} & 0.9765 & 0.8794 & 1.2052 &{0.6301} & --- & --- \\\hline
O2 & & {2.4312} & 4.9112 & 3.5057 & 3.9851 &{2.3884} & --- & --- \\\hline\hline
Locality & & {1.9713} & 2.4360 & 2.1997 & 2.1082 &{2.1227} & {1.9685} & 2.5659 \\ \hline\hline
\end{tabular}
}
             \label{tab:allcat}
\vspace{-.15in}
\end{table}

We propose several metrics to evaluate the performance of the explanation network without a human. This includes: 
\begin{enumerate}
\item {\bf Faithfulness}: We introduce a regression metric and a classification metric for faithfulness. (a) $F_{reg}=\frac{1}{M}\sum_m L({\bar y}^{(m)}-{\hat y}^{(m)})=\frac{1}{M}\sum_m|{\bar y}^{(m)}-{\hat y}^{(m)}|$, the mean absolute loss between ${\hat y^{(m)}}$ and its approximation ${\bar y^{(m)}}$; (b) We replace ${\hat y}^{(m)}$ with ${\bar y}^{(m)}$ in the original multi-class prediction vector $\hat{\bf{y}}^{(m)}$ before softmax and check whether the classification result changes. We denote $c_r$ as the number of examples whose classification results remain the same, then $F_{cls} = \frac{c_r}{M}$.
\item {\bf Orthogonality}: In order to measure whether different attention maps fall on the same region, we directly treat  attention maps of different x-features as different vectors and compute their covariance matrix. We denote ${\bf C}$ as the covariance matrix among x-features aggregated over the dataset. Then $ {\bf P} =  \text{diag}({\bf C})^{-1/2} {\bf C} \text{diag}({\bf C})^{-1/2}$ is the matrix of correlation coefficients. The orthogonality between neurons in the x-layer is defined as: (a) $O_1=||{\bf P}||_F - \sqrt{n}$, where $||\cdot||_F$ is the Frobenius norm for matrix; (b) $O_2=-\text{logdet}({\bf P})$, where $\text{logdet}$ is the logarithm of determinant of a matrix. Both $O_1$ and $O_2$ obtain the optimum at $0$, when ${\bf P}$ is a unit matrix.
\item {\bf Locality}: In order to measure locality, we propose a metric which associates x-features with various parts (CUB) or objects (Places), and measures how well they associate with these parts or objects. 
The locality for each x-feature is defined as the entropy: $\small{H_n = -\sum_p \left(\frac{{\bar S}_{p}^n}{\sum_p {\bar S}_{p}^n}\cdot\text{log}(\frac{{\bar S}_{p}^n}{\sum_p {\bar S}_{p}^n})\right)}$.
Locality is roughly measuring the log of the number of parts or objects captured by each x-feature. 
\end{enumerate}
Note that the current locality metric would not accurately represent features that do not represent a single part or object, it merely reflects our current best efforts in quantitatively measuring different explanations and especially only applicable on these limited datasets where the classification may likely to be explained by parts. Even on these datasets, we suffer from many annotated parts/objects being spatially close and semantically similar. For example, in CUB, there are $6$ parts located on the small region of the head of each bird, saliency maps usually tend to capture all these parts together hence it is very difficult to obtain a smaller number of parts on each x-feature. In places, there are a total of $173$ object categories with e.g. around $15$ different categories for various objects on a bed. Hence the locality should be interpreted more in a relative sense to compare different approaches, the absolute number of parts captured is not exactly indicative of the performance of the approach.

\begin{figure}[htb]
\begin{center}
\includegraphics[width=0.8\linewidth]{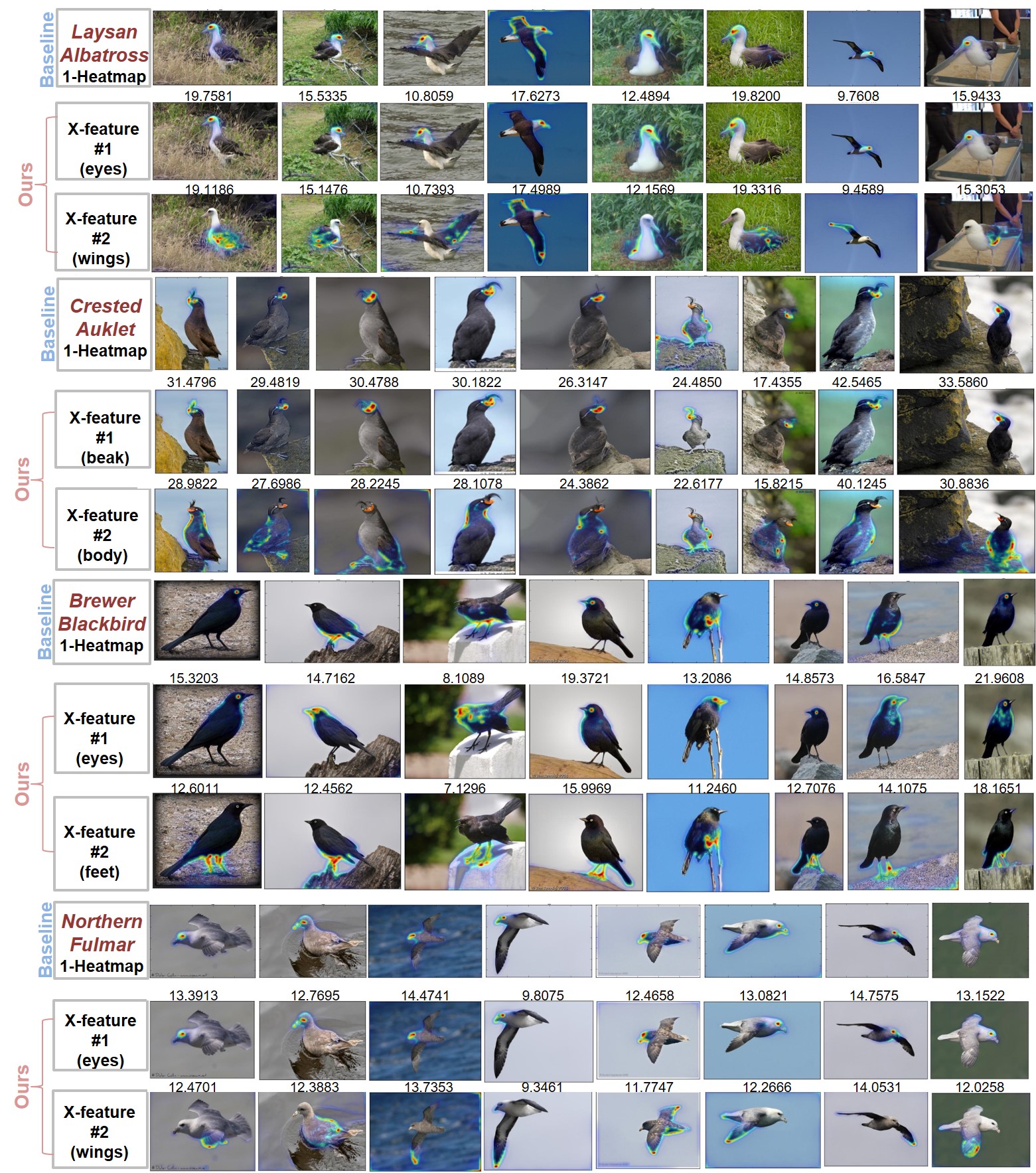}
\end{center}
\vskip -0.2in
\caption{\small Comparison of heatmaps on the original output $\hat{\bf y}$ (baseline: 1-Heatmap) and on x-features (our approach XNN) for CUB. The weight above the feature is $v_i E_i$, the product of the weight of the x-feature in the approximation of $\hat{\bf y}^{(k)}$ timed by the activation of the x-feature, which shows the contribution of the x-feature to the final prediction.}
\label{fig:CUB0}
\end{figure}

\subsection{Quantitative Results}
\newsavebox{\allcate}
\begin{lrbox}{\allcate}
\end{lrbox}
\subsubsection{Results on CUB}
 In Table \ref{tab:allcat}, we summarize the results for different explanation embedding approaches with different parameters for all the $200$ categories of the CUB dataset.  Results show that we can achieve excellent faithfulness to the predictions when using SRAE, NN, and SAE. The $F_{reg}$ in both training and testing is less than $0.2$. Since $\hat{\bf y}$ before softmax usually has a range in $[0,50]$ and especially large in the positive examples, we consider the regression loss to be small. The classification faithfulness $F_{cls}$ is even better, as only $1-2$ examples out of all the categories we tested have switched labels after replacing the original $\hat{\bf y}$ with the approximation from the x-features. This shows the major advantage of the XNN approach in that it indeed very faithfully explained the deep network. 
From Table \ref{tab:allcat}  we observe that the faithfulness for Lasso is all very bad with different parameters, indicating that it is almost impossible for the feature selection method to select a few X-features from $\bf Z$ directly to make the prediction faithful.

In terms of orthogonality and locality, our algorithm showed significant improvements over NN, SAE, and Lasso ($\alpha=2.5$ in Table \ref{tab:allcat}).
{The orthogonality of CAE is better than that of the proposed SRAE, which is reasonable because the features in ${\bf E}({\bf Z})$ are definitely more orthogonal when there is only reconstruction loss in the optimization and no attempt to achieve faithfulness. The locality of CAE is slightly worse than SRAE, but the most important problem is that it is very difficult for CAE to achieve faithfulness to the original predictions because of the lack of the faithfulness loss in the optimization.} Besides, the locality of SRAE improves significantly over the ones from 1-Heatmap, indicating that we are capable of separating information that comes from different parts. The average locality of the x-features generated by SRAE is almost matching the average locality of features in $\mathbf{Z}$. This means we are close to the limit of part separation on this layer: many of the features on the $\mathbf{Z}$ layer already represent multiple parts.

We also show some qualitative examples between heatmaps on output $\hat{y}$ (1-Heatmap) and on x-features (XNN) from different categories in Figure~\ref{fig:CUB0}. The weight above the feature is $v_i E_i$, the product of the weight of the x-feature in the approximation of $\hat{\bf y}^{(k)}$ timed by the activation of the x-feature, which shows the contribution of the x-feature to the final prediction. One can see x-features nicely separate different discriminative aspects of the bird while 1-Heatmap sometimes focuses only on one part and miss others,  and sometimes produces a heatmap that incorporates many parts simultaneously. Also, each x-feature seems distinct enough as a concept. 
More qualitative examples on the CUB dataset can be found in Figure \ref{fig:CUB1}-Figure \ref{fig:CUB4} of the Appendix.

\subsubsection{Results on Places}

\newsavebox{\allcatePlaces}
\begin{lrbox}{\allcatePlaces}
\begin{tabular}{||@{  }c@{  }|@{  }r@{  }|@{  }r@{  }|@{  }r@{  }|@{  }r@{  }|@{  }r@{  }|@{  }r@{  }|@{  }r@{  }||}
\hline
  \multicolumn{2}{||l|}{Method} &SRAE     & NN       & SAE &Lasso & {CAE}  & {\small 1-Heatmap} \\\hline\hline
$F_{reg}$       & Train &   0.5527     &     {0.3346}        &    1.4768    &  4.0726 &{4.3579}       &    ---    \\
         & Test  &  1.0260      &       0.8736        &    1.5505       & 4.3366	&{4.6553}        &    ---    \\\hline
$F_{cls}$       & Train &  97.22\%      &    97.17\%   &    94.59\%     &    90.19\% &{90.11\%}        &  ---      \\
         & Test  &   94.79\%     &     94.86\%      &    93.29\%     &   88.55\%  &{88.42\%}        &  ---      \\\hline\hline
O1       &   &  {0.2252}      &     0.3472        &   0.4578   & 0.4729   &{0.2741}          & ---       \\\hline
O2       &   & {0.5617}   &  0.8852     &     1.0799     &  0.9194  &{0.5945}        &  ---      \\\hline\hline
Locality &   & {2.7208} & 2.7756       & 2.7819  & 2.7282 &{2.7627}   & 2.7591 \\ \hline\hline
\end{tabular}
\end{lrbox}
\begin{table}[t]   
\caption{\small The average faithfulness, orthogonality, and locality of different approaches for $10$ categories of the Places dataset.} 
\vskip -0.1in
\centering   
       \scalebox{0.95}{\usebox{\allcatePlaces}}
\label{tab:Places}
\end{table}  

\begin{figure*}[t]
\begin{center}
\includegraphics[width=1\linewidth]{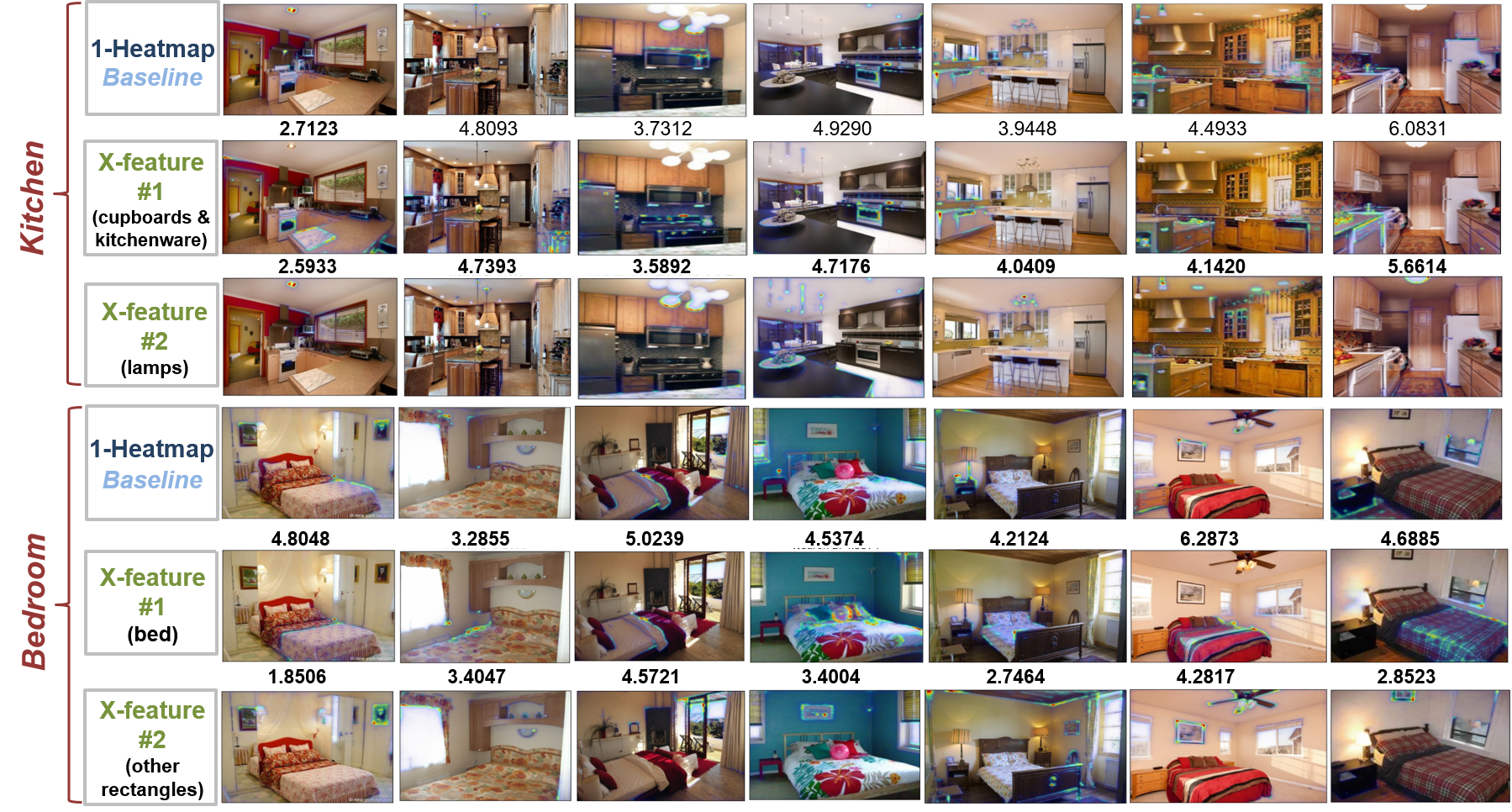}
\end{center}
\vskip -0.15in
\caption{\small Comparison of heatmaps on the original output $\hat{{\bf y}}$ (baseline: 1-Heatmap) and on x-features (our approach XNN) for {\em Kitchen} and {\em Bedroom} examples from Places dataset.}
\label{fig:exPlaces}
\end{figure*}

In Table \ref{tab:Places}, we summarize the results for different explanation embedding approaches with different parameters for $10$ categories of the Places dataset, including all the categories that represent a room and have enough images with object labels in ADE20K. 
The results on the Places dataset in Table \ref{tab:Places} show that our proposed method SRAE works well in explaining the scene recognition results for the Places dataset. 
The locality of the proposed method on Places is not as low as that on CUB, because in Places there are hundreds of different objects labels contained in ADE20K.

Figure \ref{fig:exPlaces} shows some qualitative examples between heatmaps on output $\hat{y}$ (1-Heatmap) and on x-features (XNN) for the Places dataset. The weight above the feature is $v_i E_i$, the product of the weight of the x-feature in the approximation of $\hat{\bf y}$ timed by the activation of the x-feature, which shows the contribution of the x-feature to the final prediction.

From Figure \ref{fig:exPlaces} we observe that each x-feature seems distinct enough as either a specific or a general concept.
For example, when predicting the category of {\em kitchen}, x-feature \#$1$ captures cupboards and kitchenware, and x-feature \#$2$ always focuses on lamps; while 1-Heatmap focuses on different objects in different images. 
When predicting the category of {\em bedroom}, x-feature \#$1$ always captures bed, and x-feature \#$2$ focuses on a general concept of rectangles; while 1-Heatmap still focuses on mixed concepts. The x-features surprised us as authors in both categories. In the \textit{bedroom}, we expected bed but did not expect that the \textit{rectangular corners} in picture frames and window frames would help the classification. In the \textit{kitchen}, we expected the stove being a feature, but did not expect that the CNN was actually using \textit{lighting} as a feature. These surprises showed the additional insights humans can gain of deep networks by running XNN instead of training on known concepts from language. More qualitative examples on the Places dataset can be found in Figure \ref{fig:Places1} of the Appendix. 

From the results on both CUB and Places, we believe that the x-features learned by our proposed model indeed provide concise conceptual explanations on the decisions made by CNN algorithms.

\subsection{XNN on convolutional layers}

\newsavebox{\convXNN}
\begin{lrbox}{\convXNN}
\begin{tabular}{||l|l|l|l|l|l|l||}
\hline
\multicolumn{2}{||l|}{}          & pool1 & pool2 & pool3 & pool4 & pool5 \\ \hline\hline
\multirow{2}{*}{$F_{reg}$} & Train & 2.627   & 3.530   & 4.067   & 2.605   & 2.766   \\ \cline{2-7} 
                        & Test  & 10.883  & 9.164   & 8.938   & 5.849   & 3.954   \\ \hline\hline
\end{tabular}
\end{lrbox}
\begin{table}[t]   
\caption{\small The faithfulness for the fully convolutional XNN on different convolutional blocks for the CUB dataset.} 
\vskip -0.1in
\centering   
       \scalebox{0.95}{\usebox{\convXNN}}
\label{tab:convXNN}
\end{table}  

We also performed experiments on convolutional layers to evaluate the performance of a fully convolutional XNN on the CUB dataset and the MNIST dataset. 
For CUB, the fine-tuned VGG19 model is used as the original prediction model. The explanation XNN is attached on different convolutional blocks (from pooling layer 1 to pooling layer 5),  which contains $3$ encode convolutional layers, $3$ decode convolutional layers, and a fully connected layer. 
For MNIST, the original prediction model is a simple $2$ layers CNN whose accuracy is $98.70\%$ on the testing set of MNIST. The explanation XNN is attached to the last pooling layer of the prediction model, which contains an encode convolutional layer, a decode convolutional layer, a max-pooling layer, and a fully connected layer. For the convolutional layers in XNN, the filter size is $1\times 1$. 
The number of x-features is $5$.

Table \ref{tab:convXNN} shows the faithfulness $F_{reg}$ for the convolutional XNNs from pooling layer $1$ to pooling layer $5$ on the CUB dataset. From Table \ref{tab:convXNN} we observe that the faithfulness of the convolutional layers is always significantly
worse than that of the fully connected layer. The closer it
gets to the fully connected layers, the better the faithfulness
is. And it is more likely to be overfitting for the bottom convolutional layers. Hence, we conclude that it is difficult to only utilize the features from the convolutional layers to build a fully convolutional XNN which is faithful to the prediction of the original DNN. Without the faithfulness, XNN is no longer an explanation of the original DNN. 

\begin{figure}[t]
\centering
\subfigure[] { \label{fig:convxnn}
\includegraphics[width=0.51\linewidth]{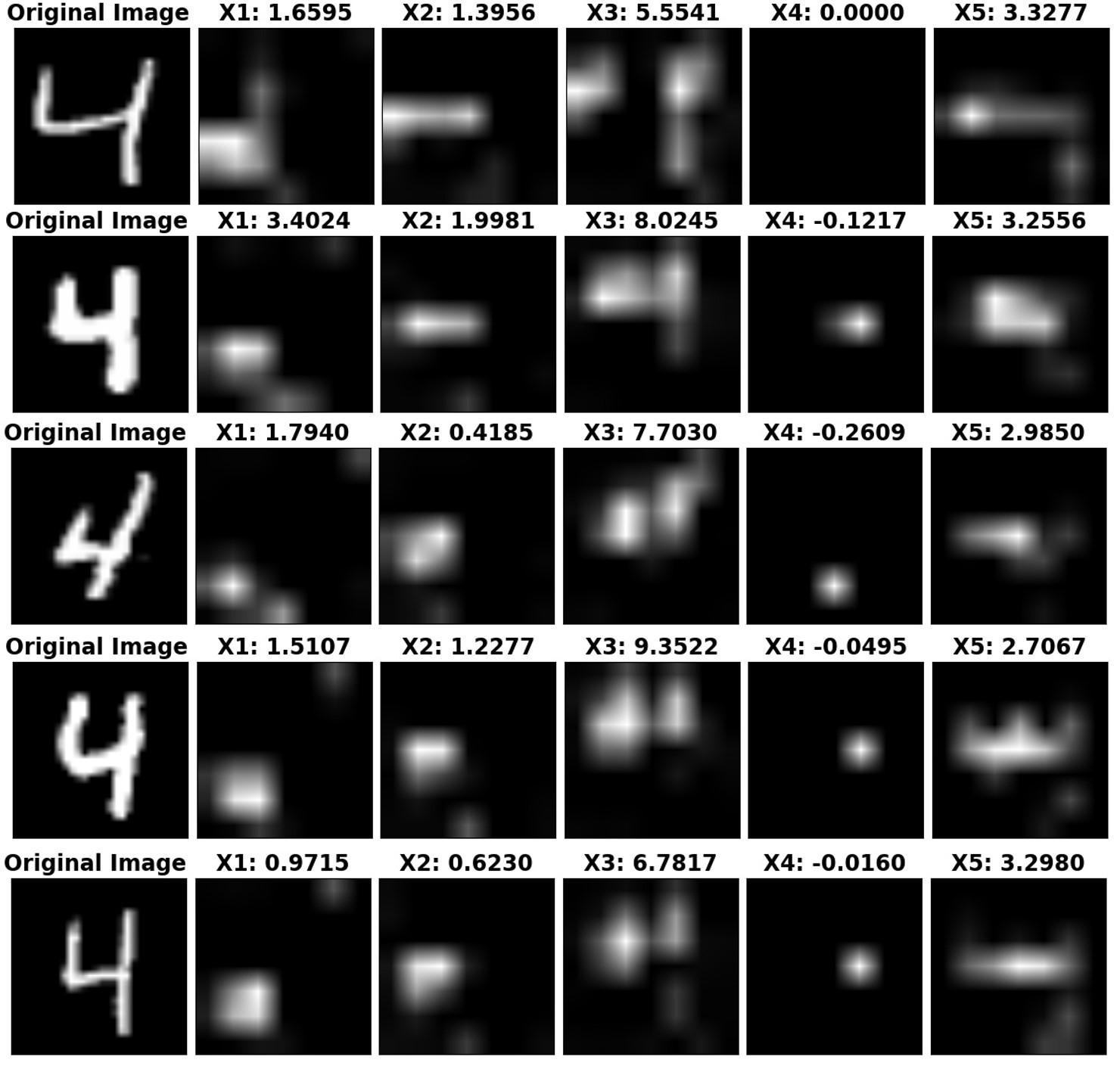}
}
\subfigure[] { \label{fig:convxnnNeg}
\includegraphics[width=0.43\linewidth]{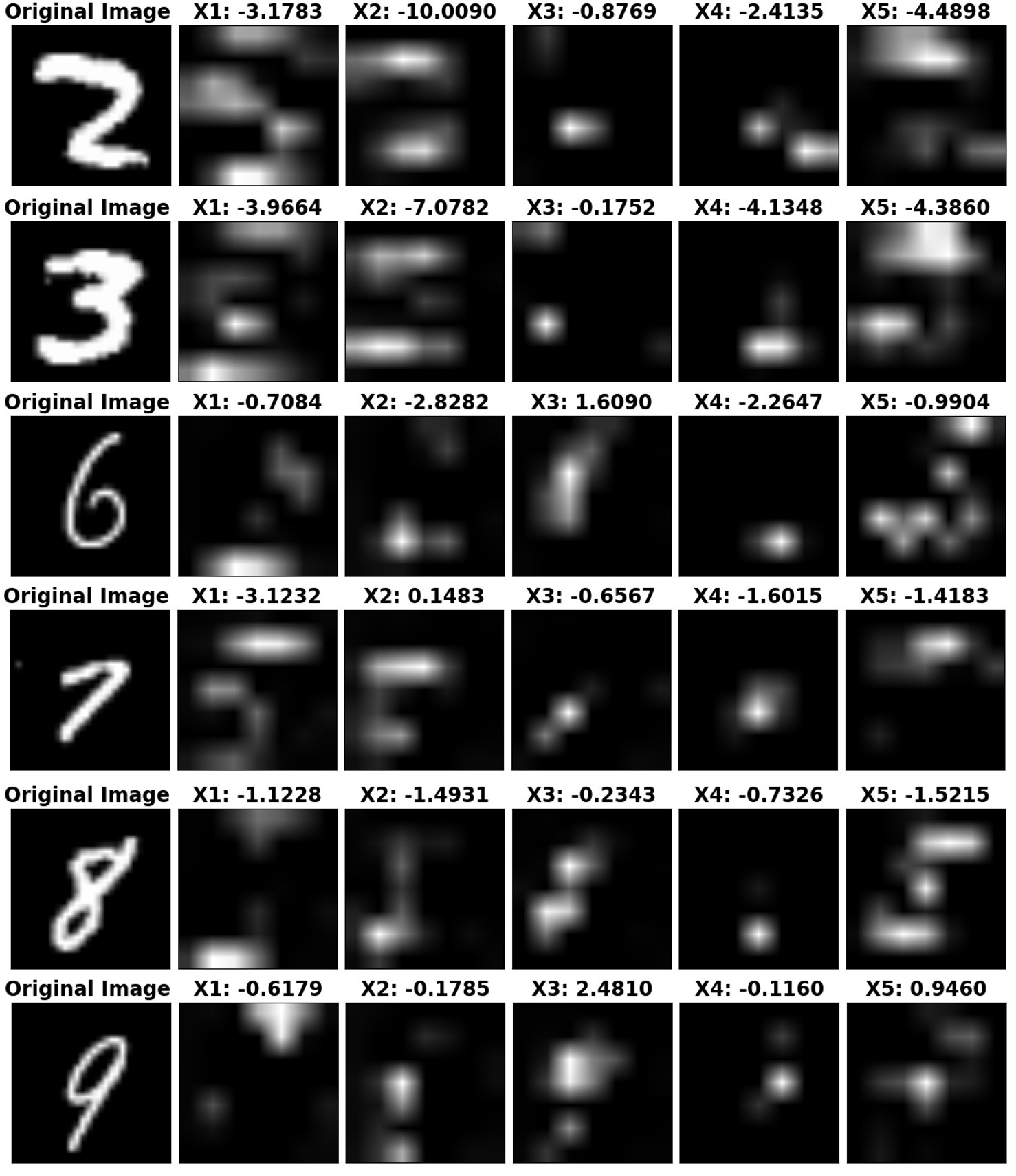}
}
\vskip -0.1in
\caption{\small (a) The visualizations of the x-features for some positive examples when explaining the category of $4$; (b) The visualizations of the x-features for some negative examples when explaining the category of $4$.}
\vskip -0.1in
\end{figure}

On MNIST, for training set, $F_{reg}=1.1952$; for testing set, $F_{reg}=2.0422$ for the convolutional XNN on the last pooling layer.
The faithfulness on MNIST is a little better than that on CUB, because the prediction model on MNIST is much simpler than that on CUB.
Figure \ref{fig:convxnn}-\ref{fig:convxnnNeg} shows some experimental results for the fully convolutional XNN on the MNIST dataset. 
Although the faithfulness on MNIST is still not as good as that of the fully connected XNN, we can find some interesting results from the x-features of the fully convolutional XNN.
Figure \ref{fig:convxnn} shows the visualizations of the x-features for some positive examples when explaining the category of number $4$; Figure \ref{fig:convxnnNeg} shows the visualizations of the x-features for some negative examples. 
From Figure \ref{fig:convxnn} we can see that the most important feature for identifying number $4$ is two vertical lines (X3); and the second important feature is one horizontal line in the middle (X5).
From Figure \ref{fig:convxnnNeg} we can see that the most irrelevant features for category $4$ are horizontal lines on the top and on the bottom.


\section{Conclusion}
In this paper, we propose an explanation network, that can be attached to any layer in a deep network to compress the layer into several concepts that can approximate an $N$-dimensional prediction output from the network. A sparse reconstruction autoencoder (SRAE) is proposed to avoid degeneracy and improve orthogonality. A human evaluation is conducted to investigate the performance of our approach against a baseline. The human study shows that on tasks that are very difficult for humans, XNN can significantly aid the human to improve its performance and understanding of deep networks. However, on tasks that are more intuitive to humans, such improvement is limited. 

We also proposed automatic evaluation metrics to evaluate the explanations on a fine-grained bird classification dataset and a scene recognition dataset. Quantitative and qualitative results show that the network can be extremely faithful to the original prediction network, and can indeed extract high-level concepts from a CNN that make sense to human, and different from existing concepts. 

There are some limitations to this work. 1) There are better heatmap approaches available now than the one utilized in the study, however at that time those newer ones are not available and we had to utilize what was the current best heatmap approach. Despite this limitation, both XNN and 1-Heatmap were using the same heatmaps, and XNN still significantly outperformed 1-Heatmap; 2) The locality metric we proposed is limited to being used on only datasets where x-features are localized, not all datasets. Also due to the uneven size of different parts and the resolution of the heatmap, the locality metric can be contaminated by localization errors. It still helped us in terms of ablation studies; 3) We performed experiments training XNN on convolutional layers, which indicated some difficulty in faithfully explaining the DNN prediction from early convolutional layers; 4) In some cases, XNN features can span multiple parts and different XNN features can exhibit some correlation, as indicated in the orthogonality and locality metrics in Table~\ref{tab:allcat}. The appendix also shows some interesting failure cases where XNN merged multiple concepts into the same x-feature.

As a research was mostly done between 2017~\cite{QiNIPSLearning17} and early 2019, XNN is our first attempt for explaining deep convolutional networks, with some successes and failures that could aid future researchers. In the future, we plan to extend XNN to explain other types of neural networks, such as recurrent networks and convolutional-recurrent ones, as well as improving the network to better separate different concepts.
We also plan to explore the relations between causality \cite{Holzinger19Cau,Holzinger16Int,Holzinger20Mea} and explainability, and try to evaluate the quality of the explanations given by an explainable AI system using the measurements defined by causality.

\section*{Acknowledgments}
This work was partially supported by DARPA contract N66001-17-2-4030.

\appendix

\section*{Appendix}

\subsection*{{\uppercase\expandafter{\romannumeral1}. Human Evaluation}}

\subsubsection*{1. Comments from participants}

In the printed questions that the participants took, they were also asked to optionally elaborate on their thoughts regarding the 2 AI agents (visualization approaches) and the overall course of the study. A few interesting comments would be referred in the following. One of the points that we noticed is that the participants found the overall task of grouping the given images difficult, especially in case of the seagulls in the CUB dataset. They were asked the question of how difficult they are finding the task of grouping images. From the answers scaled from 0 to 7, the mean value of 3.83 was obtained for the difficulty of grouping given images. 
As we indented to have a challenging task for non experts, the users elaborated this on their answers: {\em \{Not many differences between the categories!\}}, {\em \{Many of the images looked very similar, so it would be hard to parse which picture went to which group\}}, {\em \{There was a lot of similarities throughout many of the images.\}}, etc. Further, the overall performance of the participants was weaker in grouping the seagull images and this was noticed in the comments of the participants as well: {\em \{I found it very difficult to distinguish the seagulls but I found it less difficult to group the cells.\}}, {\em \{Bird images were harder [to group] than cell images.\}}, 
{\em \{Some cells had dendritic outgrowths 
that made them obvious but seagulls looked very similar.\}}, etc. This gives the intuition that since it was more plausible for the participants to find distinct features among the cell samples, they performed fairly as good as our approach in grouping the images even with no visualization. However, in the task of grouping the seagull images, as the samples were quite similar with no easy-to-find distinctive features, both the visualization methods helped the performance.

\subsubsection*{2. Participants' statistics}

In the beginning of the study, the participants were asked to select their answers for general questions about their gender, age, their background in Computer Science, etc. and  We will represent the recorded statistics from our 30 participants in the following. Age of the participants (in years): {\em \{`18-30': 76.7\%, `31-40': 16.7\%, `41-50': 3.3\%, and `51-60': 3.3\%\}}, Gender of the participants: {\em \{`Female': 43.3\%, `Male': 53.3\%, and `Trans Male': 3.3\%\}}, Background (taken courses) of the participants in Computer Science: {\em \{`No course': 36.7\%, `1 course': 36.7\%, and '4 courses or more': 26.7\%\}}, Background of the participants in Artificial Intelligence: {\em \{`No course': 76.7\%, `1 course': 3.3\%, `2 courses': 6.7\%, `3 courses': 3.3\%, and `4 courses or more': 10\%\}}.


\subsection*{{\uppercase\expandafter{\romannumeral2}. Results on the CUB dataset}}

\subsubsection*{1. Voronoi-based probability}
In Algorithm \ref{alg:Vor} we introduce the inverse distance as a factor when computing the {\bf Voronoi-based probability} $S_p^{n,m}$, trying to keep the balance between the large part region and the small part region.

\begin{algorithm}[t]
{\small
\DontPrintSemicolon
\SetKwBlock{Initialize}{Initialize}{end}
\SetKw{Compute}{Compute}
\SetKw{Update}{Update}
\SetKw{Train}{Train}
\SetKw{Let}{Let}
\SetKw{Normalize}{Normalize}
\SetKw{Select}{Select}
\SetKw{randomly}{randomly}
\SetKw{Hence}{Hence}

\ForEach{Neuron $n$ of X layer in image $I_m$}{
\ForEach{Part $p$ with its Voronoi graph $G_p$}{
\ForEach{Pixel $(i,j) \in G_p$}{
\Compute the distance between $(i,j)$ and part label $(i_p,j_p)$: $d_{ijp}=\left((i-i_p)^2+(j-j_p)^2\right)^{\frac{1}{2}}$\;
}
\Normalize the distance $d_{ijp}$ into $[0,1]$, obtain the normalized distance ${\bar d}_{ijp}$\;
\ForEach{Pixel $(i,j) \notin G_p$}{
${\bar d}_{ijp}=1$\;}
$P(\text{Part}^m_p|\text{Pixel}^m_{i,j})= 1-{\bar d}_{ijp}$\;
\Compute the probability $S_{p}^{n,m} \triangleq P(\text{Part}^m_p|\text{Neuron}_n)=\sum_{(i,j)\in I_m}P(\text{Part}^m_p|\text{Pixel}^m_{i,j}) P(\text{Pixel}^m_{i,j}|\text{Neuron}_n)=\sum_{(i,j)\in I_m}(1-{\bar d}_{ijp})S_{i,j}^{n,m}$.
}
}
\caption{{\small The metric based on Voronoi diagram}}
\label{alg:Vor}
}
\end{algorithm}
\vskip -0.05in

\subsubsection*{2. The faithfulness for Lasso}

We summarize the faithfulness for Lasso using different parameter settings for 30 randomly selected categories of the CUB dataset in Table \ref{tab:lasso}, where $\alpha$ is the parameter that multiplies the $L_1$ term in Lasso, $Num_x$ is the average number of the selected features for the 30 categories.
From Table \ref{tab:lasso} we observe that the faithfulness values for Lasso with different parameters are poor, indicating that it is almost impossible for the feature selection method to select few X-features directly from $\bf Z$ to make a faithful prediction. 

\newsavebox{\Lasso}
\begin{lrbox}{\Lasso}
\begin{tabular}{||c|r|r|r|r|r||}
\hline
Lasso &  $\alpha$ & 2.5 & 1.5 & 0.5 &0.1 \\ \hline
$Num_x$ &   & 8 & 21 & 68 & 232                \\ \hline
$F_{reg}$ &  Training & 3.80 & 3.06 & 1.86 &1.00 \\ 
            &  Testing & 3.70 & 2.99 & 1.84	 &1.03 \\ \hline\hline
\end{tabular}
\end{lrbox}
\begin{table}[t]  
\vskip -0.05in
\caption{\small The average faithfulness for Lasso with different $\alpha$ for 30 randomly selected categories of the CUB dataset.} 
\centering    
\scalebox{0.9}{\usebox{\Lasso}}
\label{tab:lasso}
\end{table}  

\subsubsection*{3. Masking the image with generated heatmaps}
To further examine whether our proposed approach offers a complete explanation of the decision made by the CNN, we attempted to do classification just using the regions that are presented in the heatmaps, similar to~\cite{gonzalez2016semantic}. 
First, we only keep top $n$ pixels that have the highest response in the heatmap, and the rest of the pixels are painted as black (we keep the highlighted regions while masking the background, which is different from \cite{gonzalez2016semantic} where the highlighted regions are removed). Then, an inpainting algorithm is applied to recover the masked images, and finally, we utilize the prediction CNN to classify the recovered images and test the classification accuracy.
Table \ref{tab:inpaint} shows results for the masking, inpainting, and classification task on 30 randomly selected categories of the CUB dataset. 
The results show that our method is slightly better than the baseline 1-Heatmap. But we would like to note that the capability of our method to generate separate concepts is non-existent in 1-Heatmap, which merges all concepts into a single heatmap. 

\newsavebox{\classifymore}
\begin{lrbox}{\classifymore}
\begin{tabular}{||c|c|c|c|c||}
\hline
 &  \tabincell{c}{Original \\image} & \tabincell{c}{Pixels \\kept} &   \tabincell{c}{Mask by \\x-features}& \tabincell{c}{Mask by \\1-Heatmap}\\ \hline\hline
\multirow{3}{0.8in}{Classification Accuracy}& \multirow{3}{0.25in}{0.8798}  & $300$ & {\bf 0.6767}             & 0.6713                \\ 
&   & $500$ & {\bf 0.7456}            &0.7381                  \\ 
 &  & $1000$ & {\bf 0.8297}            & 0.8264             \\ \hline\hline
\end{tabular}
\end{lrbox}

\begin{table}[t]   
\vskip -0.05in
\caption{\small The average classification accuracy for images masked by our method (XNN) and the baseline (1-Heatmap) with the same number of kept pixels on 30 randomly selected categories of the CUB dataset.} 
\centering    
      \scalebox{0.9}{\usebox{\classifymore}}
\label{tab:inpaint}
\end{table}  

\begin{figure*}[htbp!]
\begin{center}
\subfigure[Male downy woodpeckers] { \label{fig:Male}
\includegraphics[width=0.85\linewidth]{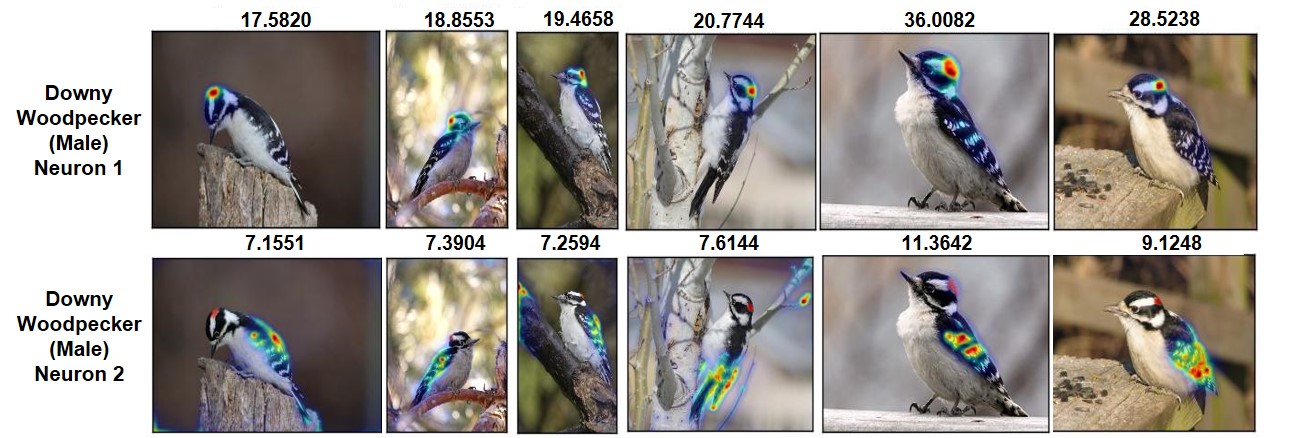}
}\\
\subfigure[Female downy woodpeckers] { \label{fig:Female}
\includegraphics[width=0.85\linewidth]{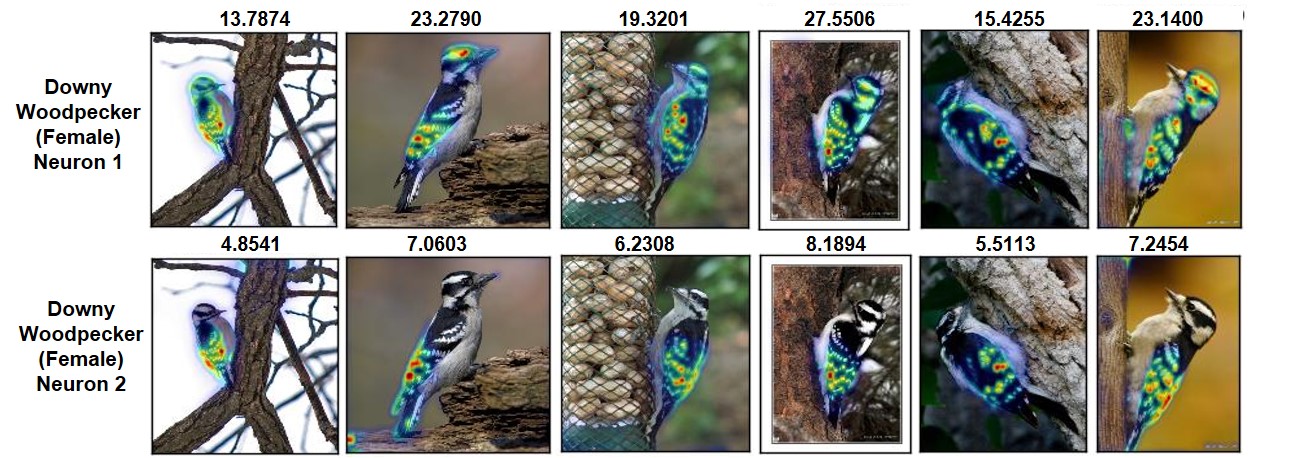}
}
\end{center}
\vspace{-1.0em}
\caption{{The x-features for male and female downy woodpeckers.}}
\vspace{-0.5em}
\label{fig:MaFe}
\end{figure*}

\subsubsection*{4. Examples of degeneration}
\begin{figure*}[]
\begin{center}
\subfigure[] { \label{fig:exgood}
\includegraphics[width=0.47\linewidth]{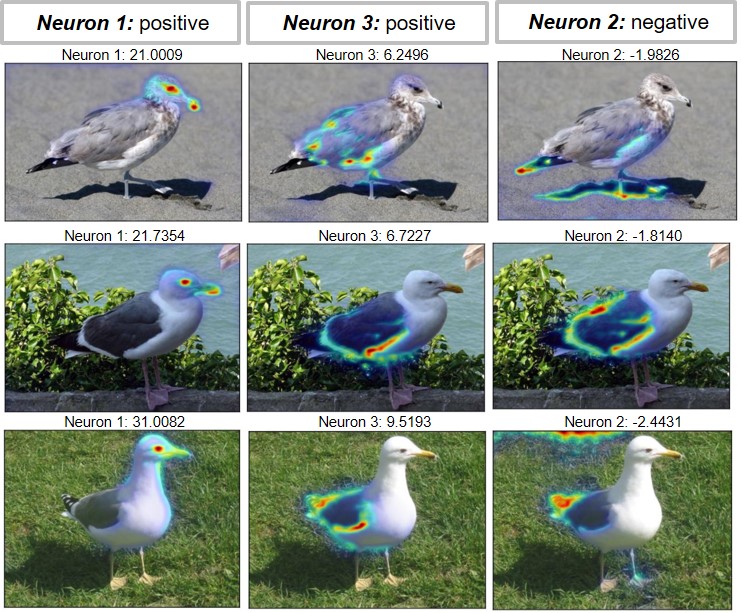}
}
\subfigure[] { \label{fig:exbad}
\includegraphics[width=0.47\linewidth]{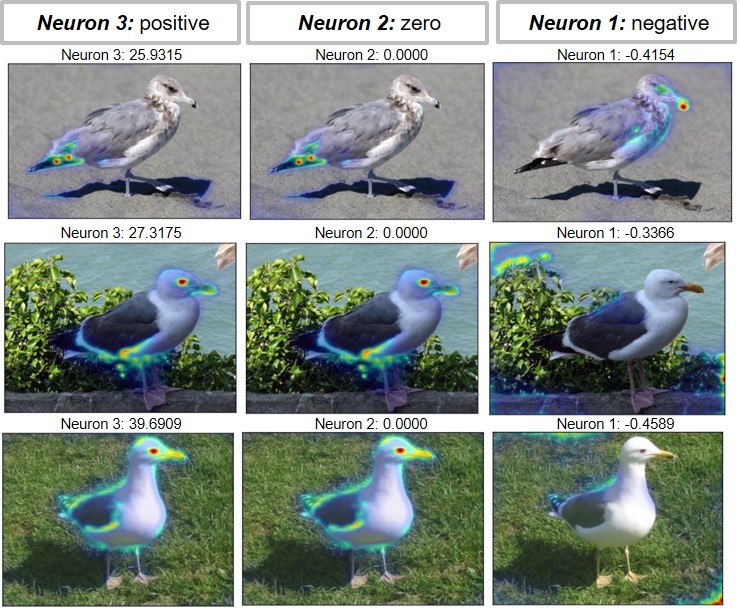}
}
\end{center}
\vspace{-1.1em}
\caption{(a) Good examples learned by SRAE, the number of the x-feature is $3$, where the $3$ neurons are orthogonal to each other; (b) Degenerated examples learned by NN, the number of the x-feature is $3$, where the first two neurons are very similar, and there is only one positive neuron.}
\vspace{-0.5em}
\label{fig:degeneration}
\end{figure*}

Figure \ref{fig:degeneration} shows some examples to illustrate the degeneration issue. Our propose method SRAE can avoid degeneration, and make the prediction model explainable.

\subsubsection*{5. An interesting failure case}

{Figure \ref{fig:MaFe} shows the x-features for male and female birds of downy woodpecker, respectively. The difference between the male and female birds of downy woodpecker is that the male birds have a red spot on the head while the female birds do not. Hence, for male birds Neuron 1 in the explanation space captures the red spot; while for female birds Neuron 1 captures the stripes on the head and the body. Neuron 2 in the explanation space captures the stripes on the body for both male and female birds of downy woodpecker. The results indicate that the x-features in the explanation space truly justify the classification decisions by capturing the key features of the birds, and the proposed model generates visualizations which are explainable to human. However, the orthogonality and locality on the female birds suffer, probably because the most indicative feature (Neuron 1) was only available in the males, hence the algorithm went on to pick some other features into Neuron 1 as well. Neuron 2 was, however, consistent in both the male and the female birds.}

\subsubsection*{6. More qualitative examples between heatmaps on original output $\hat{\bf y}$ (baseline: 1-Heatmap) and on x-features (our approach XNN)}
\begin{figure*}[t]
\begin{center}
\includegraphics[width=1\linewidth]{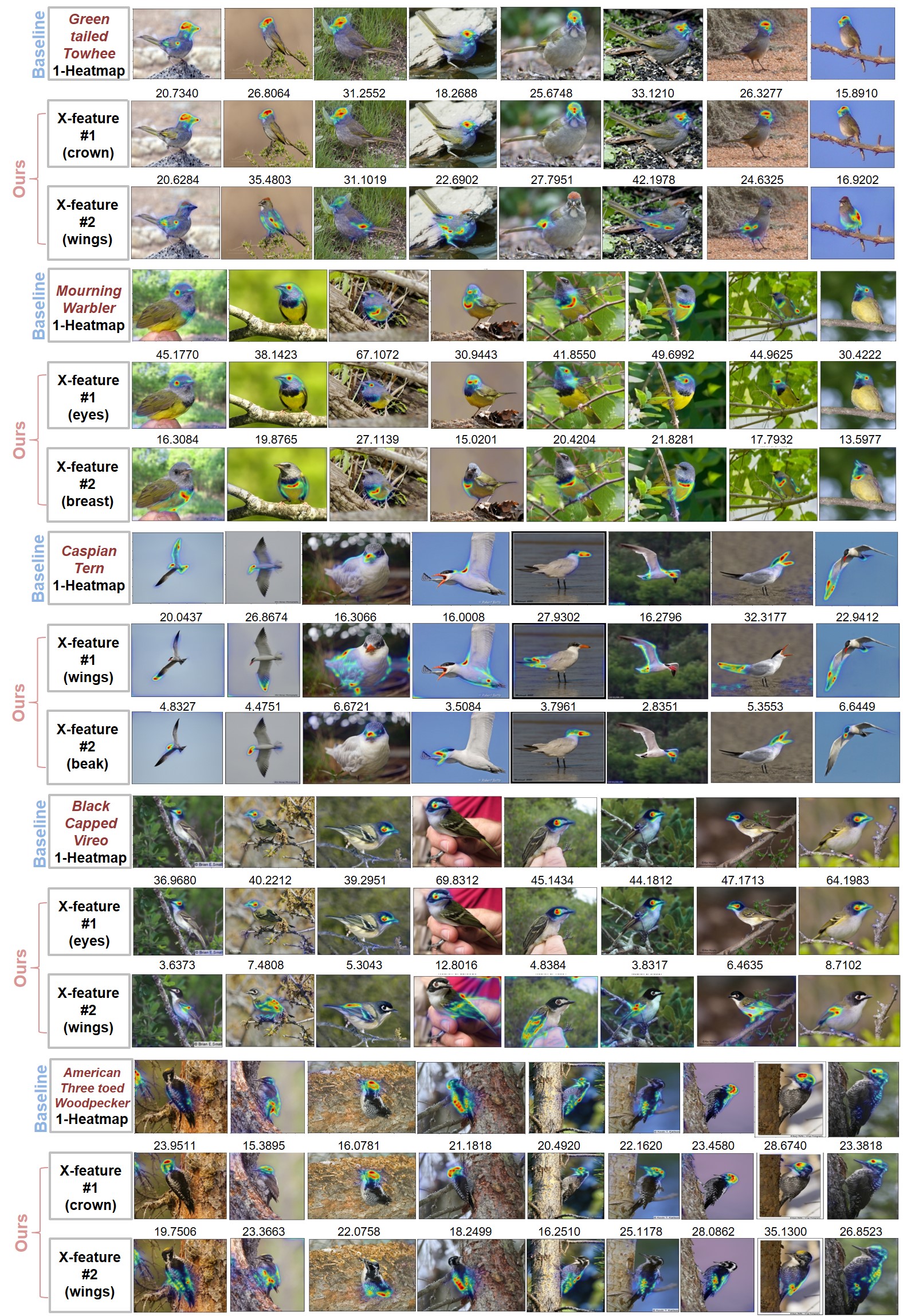}
\end{center}
\vspace{-1.0em}
\caption{Comparison of heatmaps on the original output $\hat{\bf y}$ (baseline: 1-Heatmap) and on x-features (our approach XNN) for CUB.}
\vspace{-1.0em}
\label{fig:CUB1}
\end{figure*}

\begin{figure*}[t]
\begin{center}
\includegraphics[width=1\linewidth]{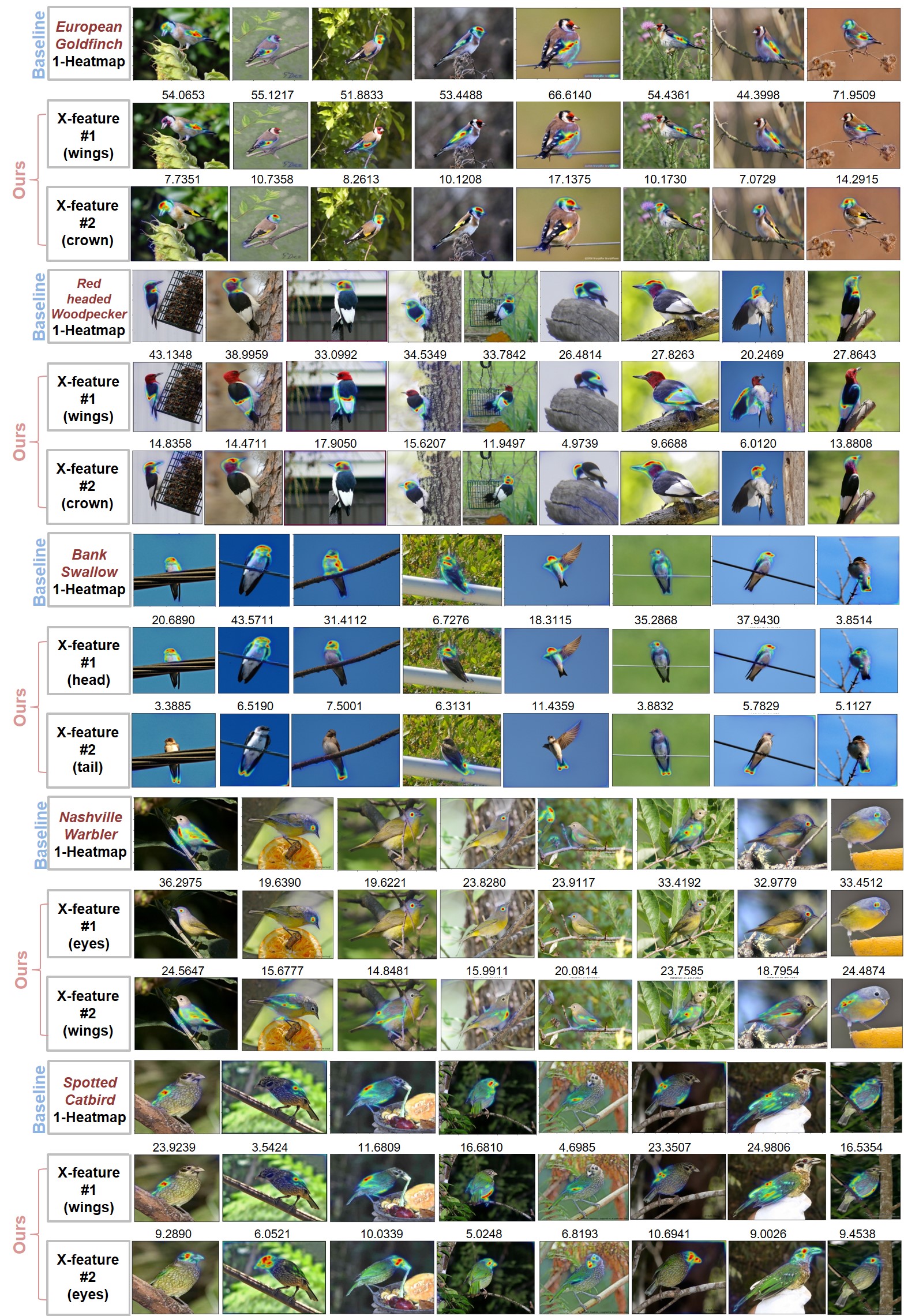}
\end{center}
\vspace{-1.0em}
\caption{Comparison of heatmaps on the original output $\hat{\bf y}$ (baseline: 1-Heatmap) and on x-features (our approach XNN) for CUB.}
\vspace{-1.0em}
\label{fig:CUB2}
\end{figure*}

\begin{figure*}[t]
\begin{center}
\includegraphics[width=1\linewidth]{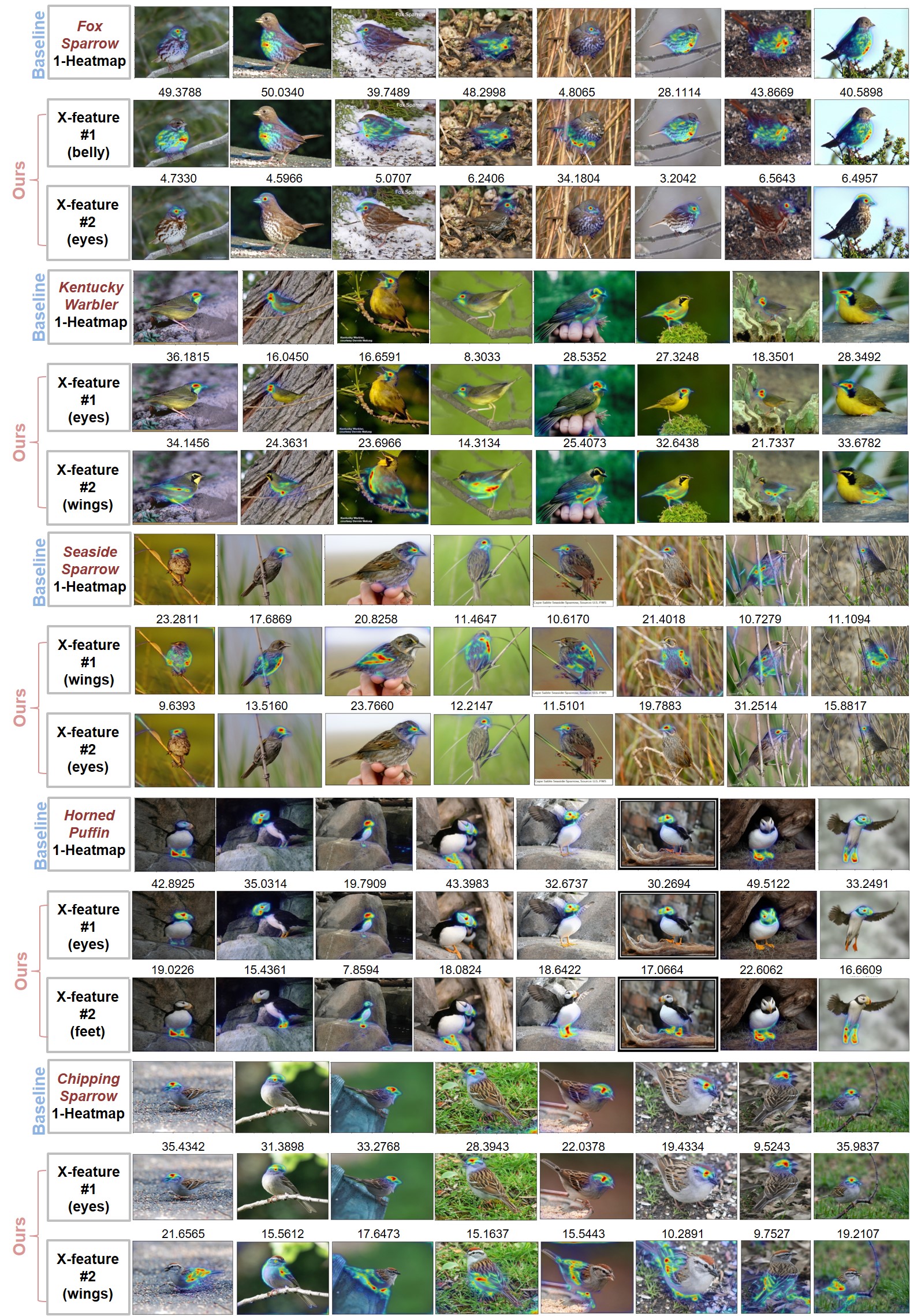}
\end{center}
\vspace{-1.0em}
\caption{Comparison of heatmaps on the original output $\hat{\bf y}$ (baseline: 1-Heatmap) and on x-features (our approach XNN) for CUB.}
\vspace{-1.0em}
\label{fig:CUB3}
\end{figure*}

\begin{figure*}[t]
\begin{center}
\includegraphics[width=1\linewidth]{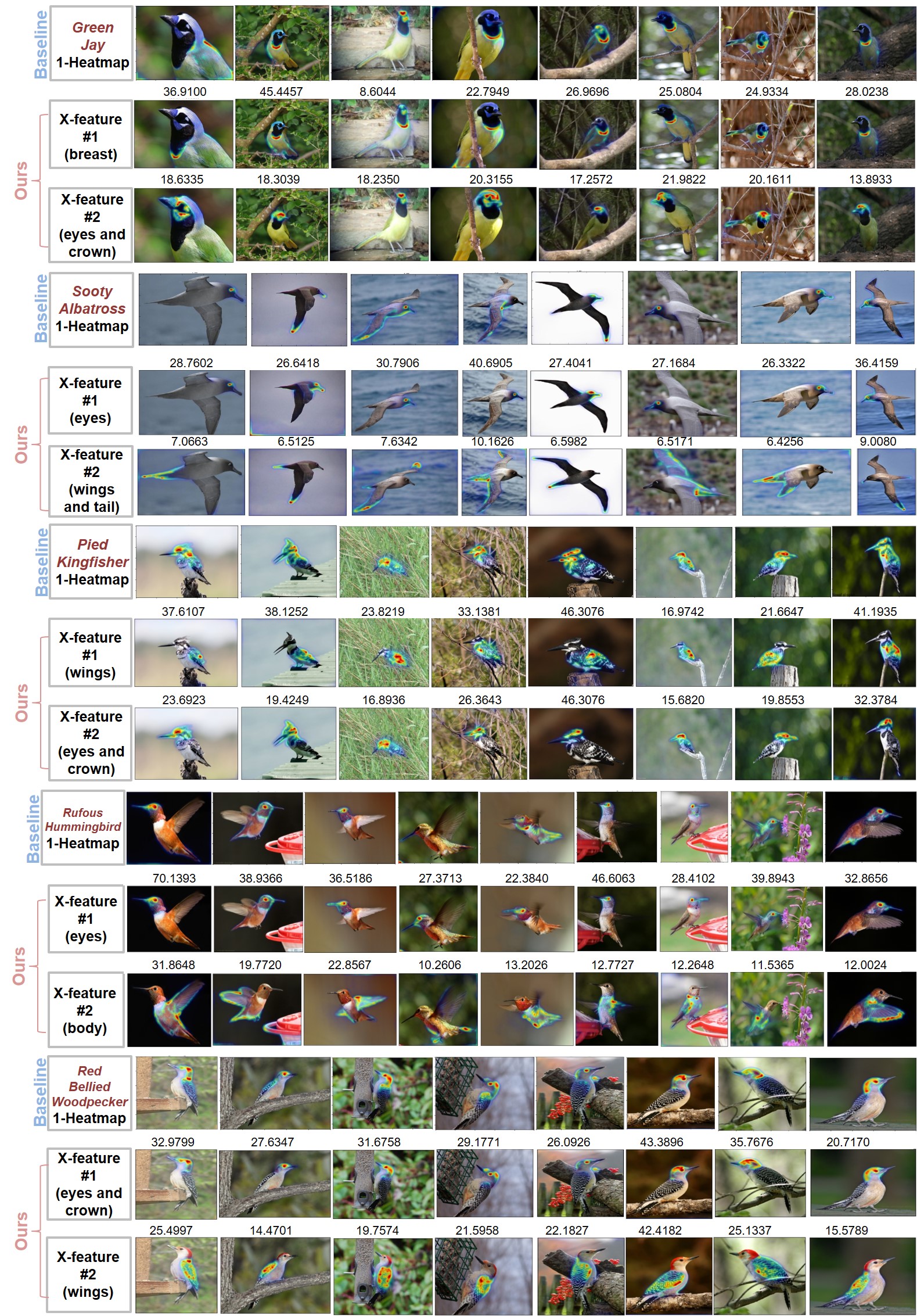}
\end{center}
\vspace{-1.0em}
\caption{Comparison of heatmaps on the original output $\hat{\bf y}$ (baseline: 1-Heatmap) and on x-features (our approach XNN) for CUB.}
\vspace{-1.0em}
\label{fig:CUB4}
\end{figure*}

Figure \ref{fig:CUB1}-Figure \ref{fig:CUB4} show more qualitative examples between heatmaps on the original output $\hat{\bf y}$ (baseline: 1-Heatmap) and on x-features (our approach XNN) for the CUB dataset.
One can see x-features nicely separate different discriminative aspects of the bird while the baseline (1-Heatmap) sometimes focuses only on one part and miss others,  and sometimes produces a heatmap that incorporates many parts simultaneously. Also, each x-feature seems distinct enough as a concept. 

\subsection*{\uppercase\expandafter{\romannumeral3}. Results on the Places dataset}

\subsubsection*{1. The faithfulness for Lasso}

For the Places dataset, we summarize the faithfulness for Lasso using different parameter settings for $10$ categories in Table \ref{tab:lassoPlaces}, where $\alpha$ is the parameter that multiplies the $L_1$ term in Lasso, $Num_x$ is the average number of the selected features for the $10$ categories.
From Table \ref{tab:lassoPlaces}, we observe that the faithfulness values for Lasso with small $Num_x$ are all very poor for different parameters, indicating that it is almost impossible for the feature selection method to select few x-features from $\bf Z$ directly to make the prediction faithful on the Places dataset.

\newsavebox{\LassoPlaces}
\begin{lrbox}{\LassoPlaces}
\begin{tabular}{||c|r|r|r|r|r|r|r||}
\hline
Lasso &  $\alpha$ & 15 & 10 & 5 &2.5&1&0.1 \\ \hline
$Num_x$ &   & 5 & 11 & 31 & 69 &283   &1782            \\ \hline
$F_{reg}$ &  Train & 4.0726 & 3.8078 & 3.3673 &3.0566 &2.2385 &1.1677\\ 
            &  Test & 4.3366 & 4.0214 & 3.5281	 &3.1655 &2.3382 &1.3131\\ \hline\hline
\end{tabular}
\end{lrbox}
\begin{table}[t]  
\vskip -0.12in
\caption{\small The average faithfulness for Lasso with different $\alpha$ on 10 categories of the Places dataset.} 
\centering    
\scalebox{0.85}{\usebox{\LassoPlaces}}
\label{tab:lassoPlaces}
\vskip -0.2in
\end{table}  

\subsubsection*{2. More qualitative examples between heatmaps on the original output $\hat{\bf y}$ (baseline: 1-Heatmap) and on x-features (our approach XNN).}
\begin{figure*}[htbp!]
\begin{center}
\includegraphics[width=1\linewidth]{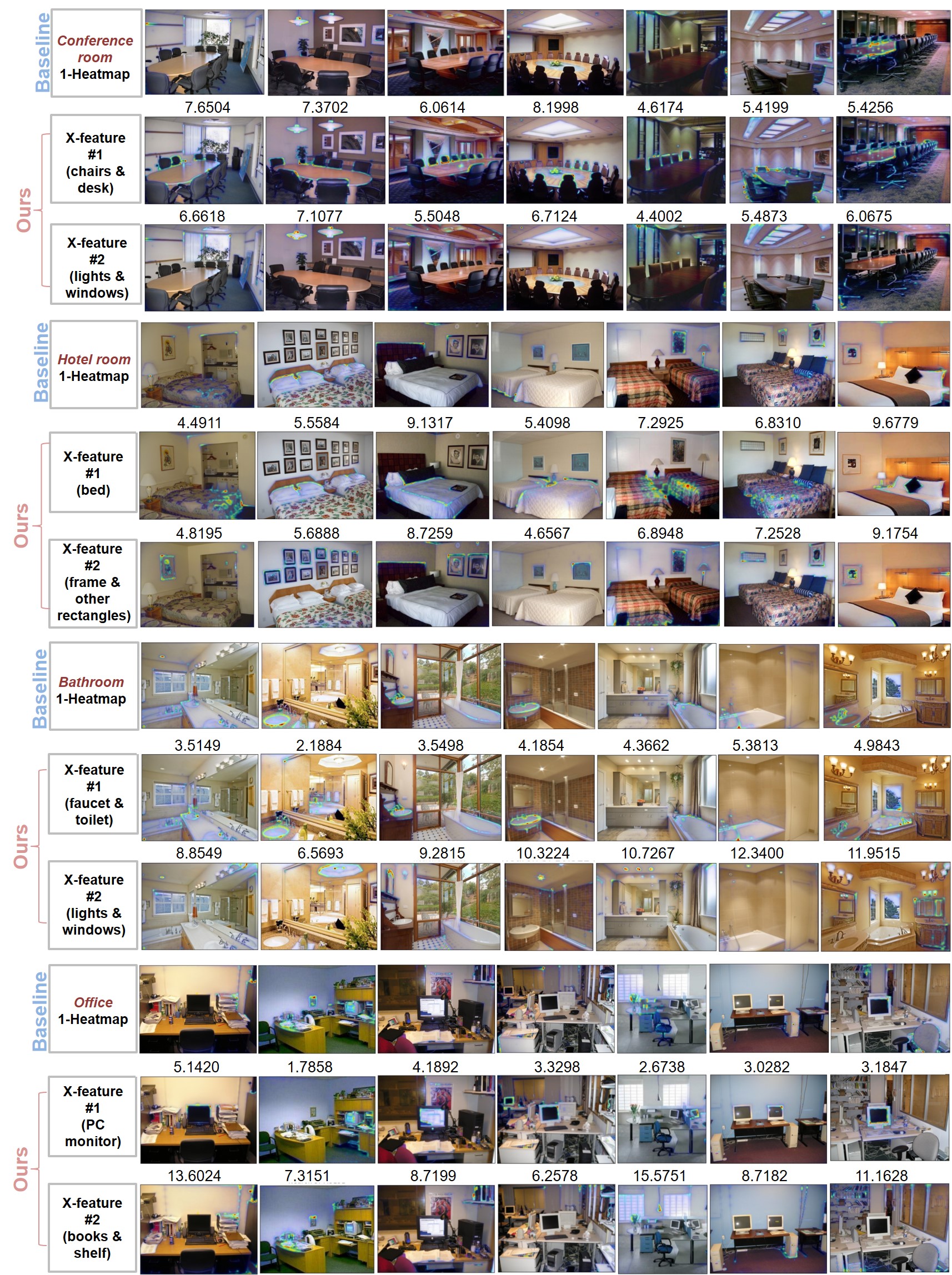}
\end{center}
\vspace{-1.0em}
\caption{Comparison of heatmaps on the original output $\hat{\bf y}$ (baseline: 1-Heatmap) and on x-features (our approach XNN) for Places.}
\vspace{-1.0em}
\label{fig:Places1}
\end{figure*}

Despite the hundreds of object labels, the x-features generated by the explanation module truly capture meaningful and consistent visual concepts on Places.
Figure \ref{fig:Places1} shows more qualitative examples between heatmaps on $\hat{\bf y}$ (baseline: 1-Heatmap) and on x-features (our approach XNN) for the Places dataset.
One can see that the x-features capture several specific or general concepts, like {\em faucet \& toilet}, {\em PC monitor}, etc.
The visualization of the baseline 1-Heatmap falls on different objects in different images, while the visualizations of x-features are more consistently focusing on the same concepts in the images.
From the results on both CUB and Places, we believe that the x-features learned by our proposed model indeed provide concise conceptual explanations of the decisions made by CNN algorithms.

\bibliographystyle{elsarticle-num}
\bibliography{egbib}

\end{document}